\documentclass[10pt,twocolumn,letterpaper]{article}

\usepackage{iccv}
\usepackage{times}
\usepackage{epsfig}
\usepackage{graphicx}
\usepackage{amsmath}
\usepackage{amssymb}

\usepackage{booktabs}       
\usepackage{amsfonts}       
\usepackage{nicefrac}       
\usepackage{microtype}      
\usepackage[export]{adjustbox}
\usepackage{algorithm,algorithmic}
\usepackage{multirow}
\usepackage{soul, xcolor}
\usepackage{caption}
\usepackage{subcaption}
\usepackage{animate}
\usepackage{array}
\usepackage{wrapfig}
\usepackage{arydshln}
\usepackage{bm}
\usepackage{mathtools}
\usepackage{bbm}
\usepackage{float}
\usepackage{xspace}
\usepackage[pagebackref=true,breaklinks=true,letterpaper=true,colorlinks,bookmarks=false]{hyperref}
\usepackage{url}
\usepackage{cleveref}
\usepackage{soul}
\usepackage[numbers]{natbib}

\usepackage{adjustbox}

\newcommand\sbullet[1][.5]{\mathbin{\vcenter{\hbox{\scalebox{#1}{$\bullet$}}}}}

\newcommand{\PreserveBackslash}[1]{\let\temp=\\#1\let\\=\temp}
\newcolumntype{C}[1]{>{\PreserveBackslash\centering}p{#1}}
\newcolumntype{R}[1]{>{\PreserveBackslash\raggedleft}p{#1}}
\newcolumntype{L}[1]{>{\PreserveBackslash\raggedright}p{#1}}

\iccvfinalcopy 

\def\iccvPaperID{3259} 

\ificcvfinal\fi

\begin{document}

\title{Strumming to the Beat: Audio-Conditioned Contrastive Video Textures}

\author{
Medhini Narasimhan$^1$ \quad Shiry Ginosar$^1$ \quad Andrew Owens$^2$ \quad Alexei Efros$^1$ \quad Trevor Darrell$^1$\\
$^1$University of California, Berkeley \quad \quad $^2$University of Michigan\\
{\tt\small \{medhini, shiry\}@berkeley.edu, ahowens@umich.edu, \{aaefros, trevordarrell\}@berkeley.edu} \\
\url{https://medhini.github.io/audio_video_textures}
}

\maketitle
\ificcvfinal\thispagestyle{empty}\fi

\begin{abstract}
We introduce a non-parametric approach for infinite video texture synthesis using a representation learned via contrastive learning. We take inspiration from Video Textures~\cite{schodl2000video}, which showed that plausible new videos could be generated from a single one by stitching its frames together in a novel yet consistent order. This classic work, however, was constrained by its use of hand-designed distance metrics, limiting its use to simple, repetitive videos. We draw on recent techniques from self-supervised learning to learn this distance metric, allowing us to compare frames in a manner that scales to more challenging dynamics, and to condition on other data, such as audio. We learn representations for video frames and frame-to-frame transition probabilities by fitting a video-specific model trained using contrastive learning. To synthesize a texture, we randomly sample frames with high transition probabilities to generate diverse temporally smooth videos with novel sequences and transitions. The model naturally extends to an audio-conditioned setting without requiring any finetuning. Our model outperforms baselines on human perceptual scores, can handle a diverse range of input videos, and can combine semantic and audio-visual cues in order to synthesize videos that synchronize well with an audio signal. 
\end{abstract}

\section{Introduction}

We revisit Video Textures~\citep{schodl2000video}, a classic non-parametric video synthesis method  which converts a single input video into an infinitely long and continuously varying video sequence. Video textures have been used to create dynamic backdrops for special effects and games, 3D portraits, dynamic scenes on web pages, and the interactive control of video-based animation~\cite{schodl2001machine,schodl2002controlled}. In these models, a new plausible video texture is generated by stitching together snippets of an existing video. Classic video texture methods have been very successful on simple videos with a high degree of regularity, such as a swinging pendulum. However, their reliance on Euclidean pixel distance as a similarity metric between frames makes them brittle to irregularities and chaotic movements, such as dances or performance of a musical instrument. They are also sensitive to subtle changes in brightness and often produce jarring transitions.

\begin{figure}[t]
    \centering
    \begin{adjustbox}{max width=\linewidth}
    \setlength{\tabcolsep}{1pt}
    \begin{tabular}{cc}
        \href{https://www.youtube.com/watch?v=YFSYgLAAoXE}{\includegraphics[width=0.4\linewidth]{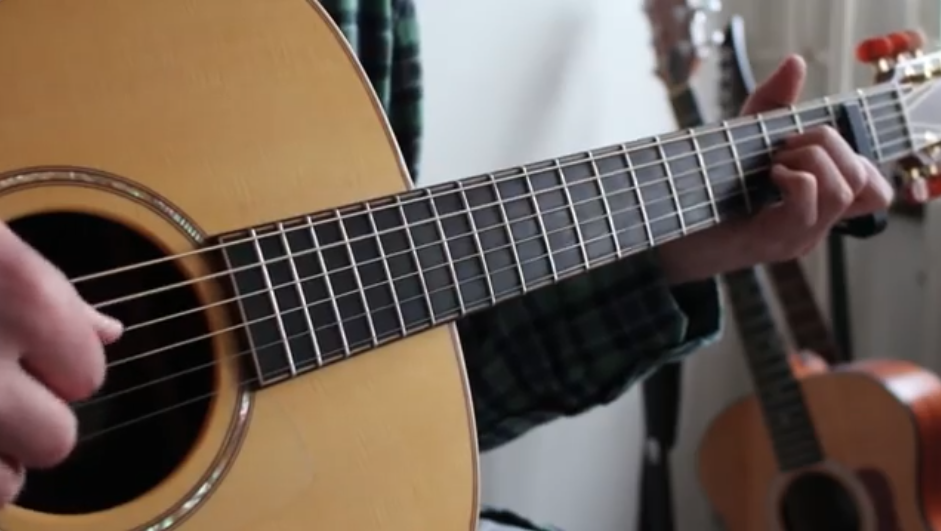}} & \href{https://medhini.github.io/audio_video_textures/arxiv_videos/morningsun.wav}{\includegraphics[width=0.4\linewidth] {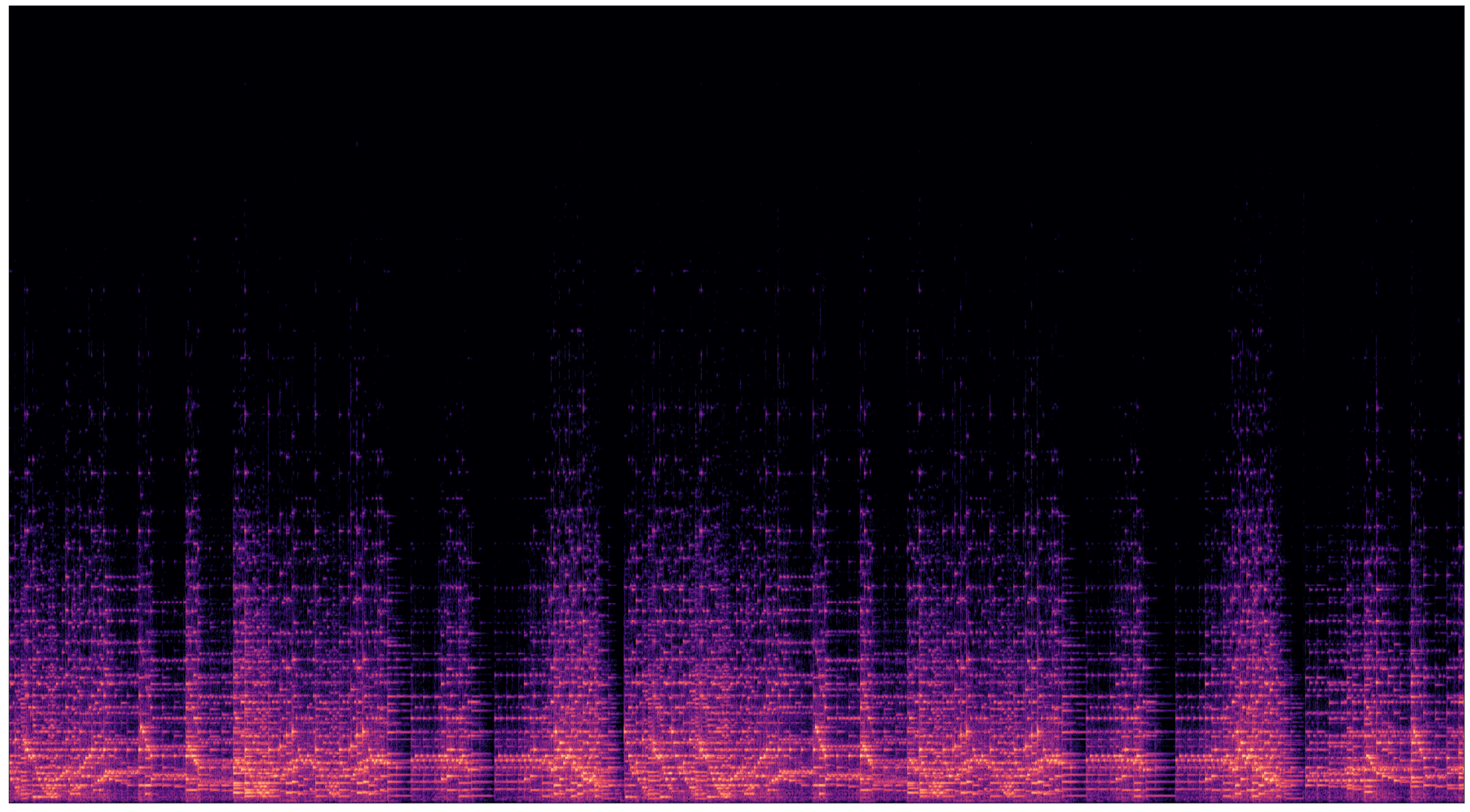}}\\
        {Input Video} & {Conditioning Audio} \\
        \multicolumn{2}{c}{\includegraphics[width=0.8\linewidth]{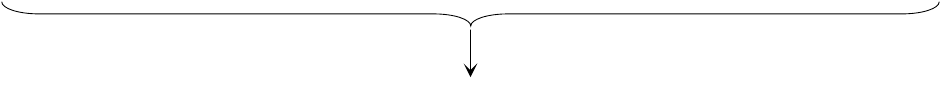}} \vspace{-0.2cm}\\
        \multicolumn{2}{c}{\href{https://medhini.github.io/audio_video_textures/arxiv_videos/sadguitar_morningsun.mp4}{\includegraphics[width=0.8\linewidth]{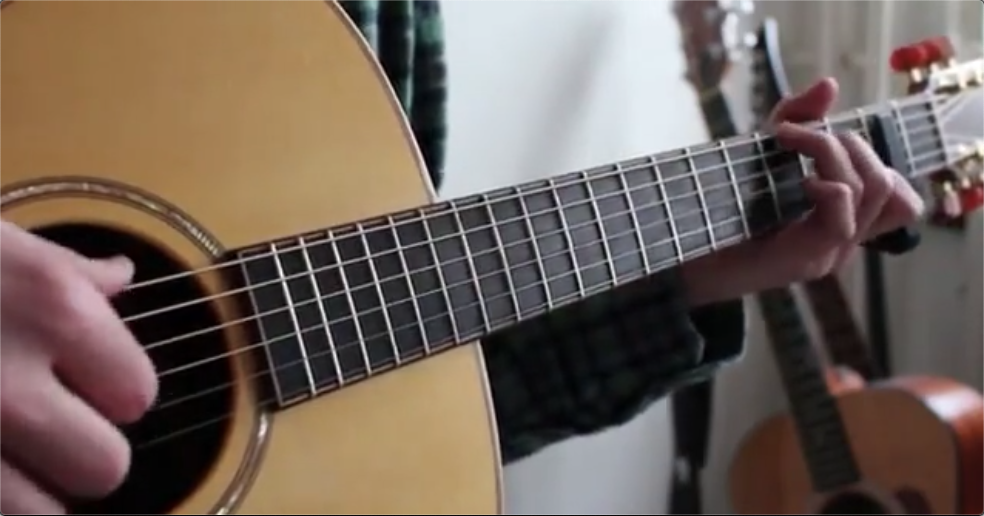}}} \\
        \multicolumn{2}{c}{Audio-Conditioned Contrastive Video Texture}
    \end{tabular}
    \end{adjustbox}
    \captionof{figure}{\textbf{Strumming to the Beat.} Click on each image to play the video/audio. We introduce Contrastive Video Textures, a learning-based approach for video texture synthesis. Given an input video and a conditioning audio, we extend our Contrastive model to synthesize a video texture that matches the conditioning audio. }
    \label{fig:teaser1}
\end{figure}

\begin{figure*}[t]
    \centering
    \includegraphics[width=\linewidth]{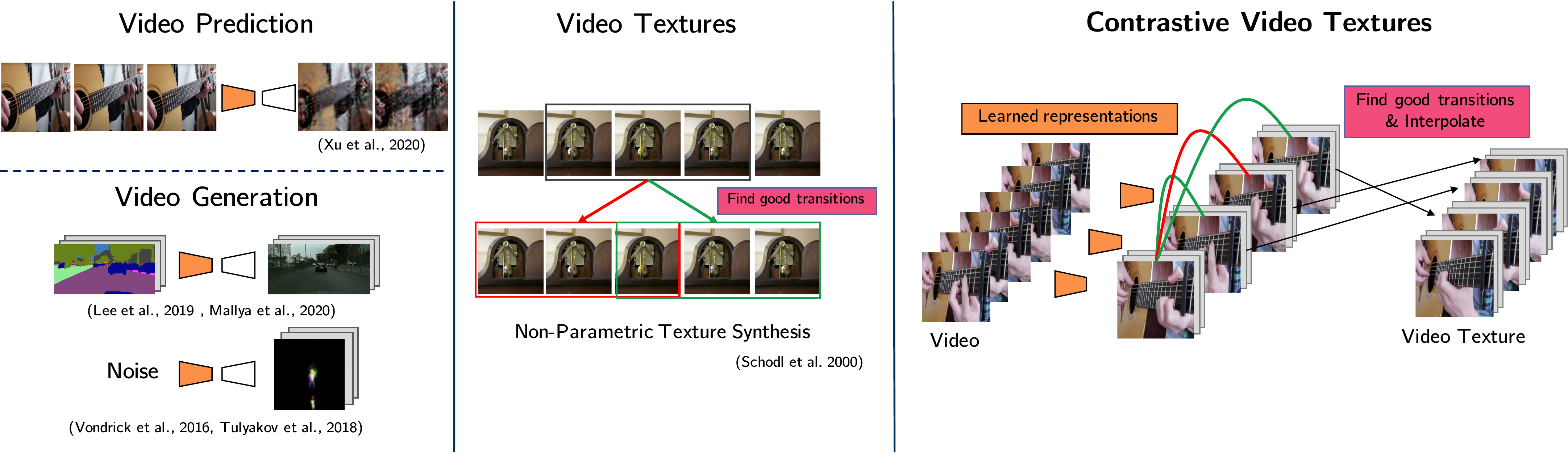}
    \captionof{figure}{\textbf{Video Texture Synthesis.} Prior video prediction~\citep{xu2020video} and generation~\citep{ tulyakov2018mocogan, vondrick2016generating} methods fail to generate long and diverse video textures at a high resolution. Vid2Vid~\citep{lee2019dancing,mallya2020world} methods require semantic maps as input and aren't suitable for video texture synthesis. Classic video textures~\citep{schodl2000video} (middle) can generate infinite sequences by resampling frames, but uses fixed representations which are not robust to varying domains. Our method (right) learns a representation and non-parametric method for infinite video texture synthesis based on resampling frames from an input video. }
    \label{fig:teaser2}
\end{figure*}

Representation-learning methods have made significant advances in the past decade and offer a potential solution to the limitations of classic video texture approaches. A natural approach may be to use 
Generative Adversarial Networks (GANs)~\citep{goodfellow2014generative} and/or Variational Autoencoders (VAEs)~\citep{kingma2013auto} which have achieved great success in generating images ``from scratch''. Yet while video generation~\citep{lee2019dancing,mallya2020world,tulyakov2018mocogan,vondrick2016generating,wang2018vid2vid, wang2018video} has shown some success, videos produced using such methods are unable to match the realism of actual videos. Current generative video methods fail to capture the typical temporal dynamics of real video and as a result fail on our task of synthesizing long and diverse video sequences conditioned on a single source video. In this work, we investigate contrastive learning~\citep{chen2020simple,chen2020improved,chen2019mocycle} approaches to graph-based sequence generation, conditional and unconditional, and demonstrate the ability of learned visual texture representations to render compelling video textures. 

We propose Contrastive Video Textures, a non-parametric learning-based approach for video texture synthesis that overcomes the aforementioned limitations. As in~\cite{schodl2000video}, we synthesize textures by resampling frames from the input video. However, as opposed to using pixel similarity, we {\em learn} feature representations and a distance metric to compare frames by training a deep model on \emph{a single input video}. The network is trained using contrastive learning to fit an example-specific bi-gram model ({\em i.e.} a Markov chain). This allows us to learn features that are spatially and temporally best suited to the input video.

To synthesize the video texture, we use the video-specific model to compute probabilities of transitioning between frames of the same video. We represent the video as a graph where the individual frames are nodes and the edges represent transition probabilities predicted by our video-specific model. We generate output videos (or textures) by randomly traversing edges with high transition probabilities. Our proposed method is able to synthesize realistic, smooth, and diverse output textures on a variety of dance and music videos as shown at this \href{https://medhini.github.io/audio_video_textures/}{\underline{website}}. Fig.~\ref{fig:teaser2} illustrates the distinction between video generation/prediction, video textures, and our contrastive model.

Learning the feature representations allows us to easily extend our model to an audio-conditioned video synthesis task. A demonstration of this task where the guitarist is ``strumming to the beats" of a new song is shown in Fig.~\ref{fig:teaser1}. Given a source video with associated audio and a new {\em conditioning} audio not in the source, we synthesize a new video that matches the conditioning audio. We modify the inference algorithm to include an additional constraint that the predicted frame's audio should match the conditioning audio. We trade off between temporal coherence (frames predicted by the constrastive video texture model) and audio similarity (frames predicted by the audio matching algorithm) to generate videos that are temporally smooth and also align well with the conditioning audio. 

We assess the quality of the synthesized textures by conducting human perceptual evaluations comparing our method to a number of baselines. In the case of unconditional video texture synthesis, we compare to the classic video texture algorithm~\citep{schodl2000video} and variations to this which we describe in Sec.~\ref{sec:exp}. For the audio-conditioning setting, we compare to four different baselines: classic video textures with audio-conditioning, visual rhythm and beat~\citep{davis2018visual}, Audio Nearest-Neighbours, and a random baseline. Our studies confirm that our method is perceptually better than all of these previous methods.
\section{Related Work}

\begin{figure*}[t]
    \centering
    \includegraphics[width=\textwidth]{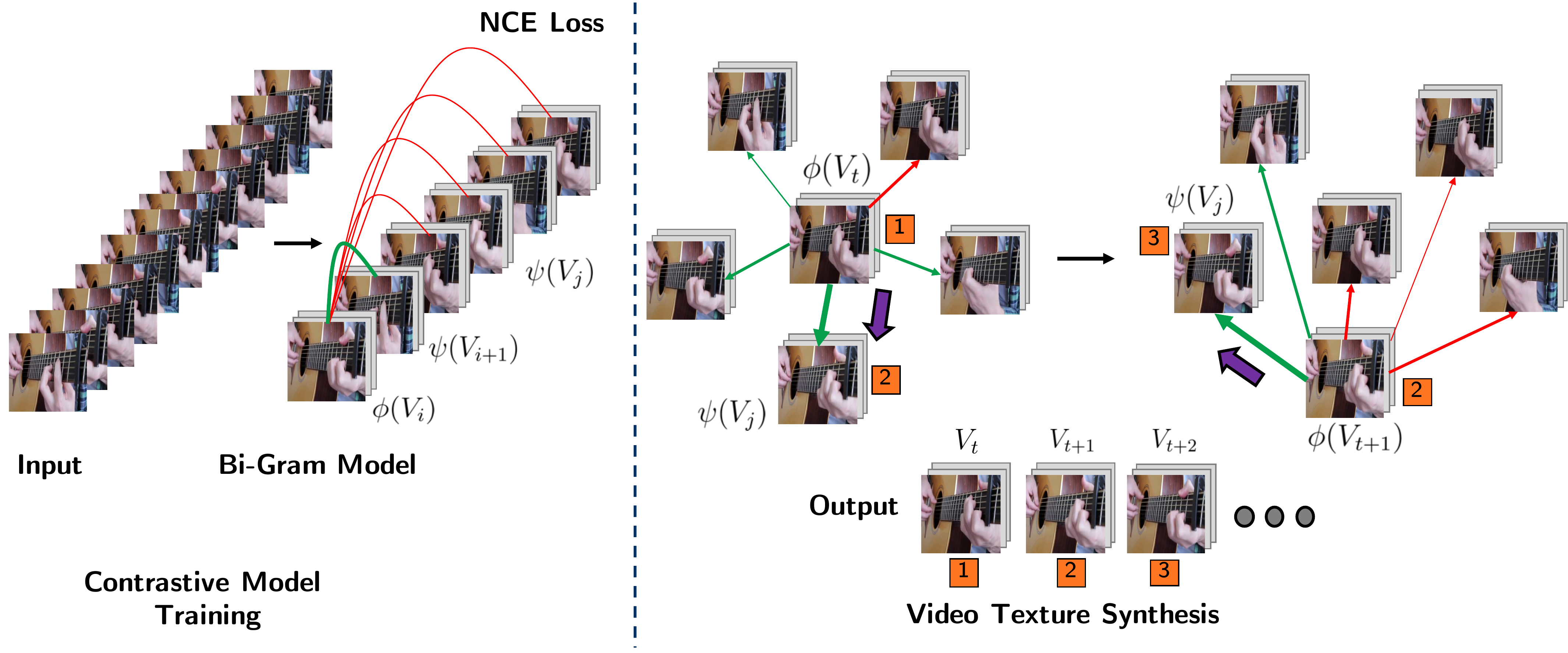}
    \caption{\textbf{Contrastive Video Textures.} We extract overlapping segments from the video and fit a bi-gram model trained using NCE loss (Eq.~\ref{eq:likelihood}) which learns representations for query/target pairs such that given a query segment $V_i$, $\phi(V_i)$ is similar to positive segment $\psi(V_{i+1})$ and dissimilar to negative segments $\psi(V_j)$ where $j \in [1,... N]$ and $j \neq i, i+1$. \textbf{Video Texture Synthesis.} During inference, we start with a random segment $V_t$ shown by \protect\includegraphics[height=0.3cm]{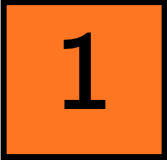}, compute $\phi(V_t)$ and $\psi(V_j) ~\forall~j \in [1, ...N]$ and calculate the edge weights as similarity between $\phi(V_t)$ and $\psi(V_j)$. We denote higher weight edges in \textcolor{green}{green} and lower weighted edges in \textcolor{red}{red} and the thickness correlates with the probability. We randomly traverse (\textcolor{violet}{purple} arrow) along one of the higher weighted edges to reach \protect\includegraphics[height=0.3cm]{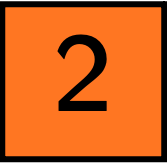}.  \protect\includegraphics[height=0.3cm]{figs/1.png} and \protect\includegraphics[height=0.3cm]{figs/2.png} are appended to the output and the process is repeated with \protect\includegraphics[height=0.3cm]{figs/2.png} as the query.} 
    \label{fig:approach}
\end{figure*}

\noindent\textbf{Texture Synthesis.} All texture synthesis methods aim to produce textures which are sufficiently different from the source yet appear to be produced by the same underlying stochastic process. Texture synthesis methods can be broadly classified into two categories: non-parametric and parametric. Non-parametric methods focus on modeling the conditional distribution of the input images and sample information directly from the input. The sampling could be done pixel-wise~\citep{efros1999texture, wei2000fast} or patch-wise~\citep{efros2001image, kwatra2003graphcut} for image texture synthesis. Wei~\etal~\citep{wei2009state} provides an extensive review of example-based texture synthesis methods. Parametric approaches, on the other hand, focus on explicitly modeling the underlying texture synthesis process. Heeger~\etal~\citep{heeger1995pyramid} and Portilla~\etal~\cite{portilla2000parametric} were the first to propose parametric image texture synthesis by matching statistics of image features between source and target images. This later inspired Gatys~\etal~\citep{gatys2015texture}, which used features learned using a convolutional neural network for image texture synthesis. Inspired by these works,~\citep{schodl2000video} proposed a non-paramteric approach for synthesizing a video texture with by finding novel, plausible transitions in an input video. Following work~\citep{efros2003,schodl2001machine, schodl2002controlled} explored interesting extensions of the same. All these video texture synthesis works use Euclidean pixel distance as a similarity measure. This causes the texture synthesis to fail on more complex scenes. On the other hand, our learned contrastive feature  representations and similarity metric generalizes well to dance/music domains and also allows for conditioning on heterogeneous data such as audio. 

\medskip
\noindent\textbf{Video Generation and Video Prediction.}  The success of (GANs)~\citep{goodfellow2014generative} and Variational Autoencoders (VAEs)~\citep{kingma2013auto} in image generation~\citep{ karras2018style,park2019semantic,zhu2017unpaired} inspired several video generation methods, both unconditional~\citep{clark2019efficient,holynski2020animating,saito2017temporal,tulyakov2018mocogan,vondrick2016generating} and conditional~\citep{chen2019mocycle, gafni2019vid2game,mallya2020world,menapace2021playable,wang2019few,wang2018video,zhang2020vid2player,zhou2019dance}. 
While conditional video synthesis of future frame prediction given past frames~\citep{denton2017unsupervised,kalchbrenner2016video,srivastava2015unsupervised,xu2020video,ye2019compositional} works well, these methods are far from generating realistic infinitely-long and diverse video. They oftentimes produce outputs which are low-resolution, especially in the unconditional case. This is because videos are higher dimensional and modeling spatio-temporal changes and transition dynamics is more complex. As such, these methods are expected to fail when applied to our task of video texture synthesis which involves rendering a video as an infinitely varying stream of images. This requires capturing the temporal dynamics of the video which current video generation methods fail to do. Similar to recent works which condition the video generation on an input signal such as text~\citep{li2018video}, or speech~\citep{ephrat2018looking,kim2018learning,oh2019speech2face}, or a single image~\citep{shaham2019singan}, we condition video texture synthesis on an audio signal. Our work is inspired by test-time training methods such as SinGAN~\citep{shaham2019singan}, Deep Image Prior~\citep{ulyanov2018deep}, and Patch VAE-GAN~\citep{gur2020hierarchical} in that we train an example-specific model on a single input, though on a video instead of an image and without an adversarial loss. Our method only takes a few hours to train on a single video and doesn't require hours of training on a large dataset. 

\medskip
\noindent\textbf{Contrastive Learning.} Recent contrastive learning approaches~\citep{chen2020simple,chen2020improved,chen2019mocycle, he2020momentum,henaff2019data} have achieved success in classic vision tasks proving the usefulness of the learned representations. Mishra~\etal~\citep{misra2016shuffle} train a network to determine the temporal ordering of frames in a video and Wei~\etal~\citep{wei2018learning}'s self-supervised model learns to tell if a video is playing forwards/backwards. Here, we use contrastive learning to fit a video-specific bi-gram model. Our network maximizes similarity between learned representations for the current and next frame. Unlike ~\citep{oord2018representation}, our goal is not to generate frames from latent representations, but rather to use the learned distance metric to resample from the input video.    

\section{Contrastive Video Textures}
\label{sec:cvt}

An overview of our method is provided in Fig.~\ref{fig:approach}. We propose a non-parametric learning-based approach for video texture synthesis. At a high-level, we fit an example-specific bi-gram model (\ie a Markov chain) and use it to re-sample input frames, producing a diverse and temporally coherent video. In the following, we first define the bi-gram model, and then describe how to train and sample from it.

Given an input video, we extract $N$ overlapping segments denoted by $V_{i}$ where $i \in [1,... N]$, with a sliding window of length $W$ and stride $s$. Consider these segments to be the states of a Markov chain, where the probability of transition is computed by a deep similarity function parameterized by encoders $\phi$ and $\psi$:
\begin{equation} 
    P(V_{i+1} | V_{i}) \propto \exp(\mathrm{sim}(\phi(V_i), \psi(V_{i+1}))/\tau)
    \label{eq:video_enc}
\end{equation}

We use two separate encoder heads $\phi$ and $\psi$ for the query and target, respectively, to break the symmetry between the two embeddings. This ensures $\mathrm{sim}(V_i, V_{i+1}) \neq \mathrm{sim}(V_{i+1}, V_i)$, which allows the model to learn the arrow of time. Fitting the transition probabilities amounts to fitting the parameters of $\phi$ and $\psi$, which here will take form of a 3D convolutional network. The model is trained using temperature-scaled and normalized NCE Loss~\citep{mikolov2013distributed}: 
\begin{align}
    \mathcal{L}(V, \phi) &= \sum_{i=1}^N -\log P(V_{i+1} | V_i) \nonumber \\
        &= \sum_{i=1}^N -\textrm{log}\frac{\exp(S(V_i, V_{i+1})/\tau)}{{\sum_{j=1}^{N}} \mathbbm{1}_{[j \notin \{i, i+1\}]} \exp(S(V_i, V_j)/\tau)} \nonumber \\
        &\textrm{where, } S(V_i, V_j) = \mathrm{sim}(\phi(V_i), \psi(V_{j})) 
    \label{eq:likelihood}
\end{align}
where $\tau$ denotes a temperature term that modulates the sharpness of the softmax distribution. As the complexity increases with number of negatives in the denominator, for efficiency, we use negative sampling~\citep{mikolov2013distributed} to approximate the denominator in Eq~\ref{eq:likelihood}. Fitting the encoder in this manner amounts to learning a video representation by contrastive learning, where the positive is the segment that follows, and negatives are sampled from the set of all other segments. The encoder thus learns features useful for predicting the dynamics of phenomena specific to the input video. 
 
\medskip
\noindent\textbf{Video Texture Synthesis.} To synthesize the texture, we represent the video as a graph, with nodes as segments and edges indicating the transition probabilities computed by our Contrastive model as shown in Fig.~\ref{fig:approach}. We randomly select a query segment $V_t$ among the segments of the video and set the output sequence to all the $W$ frames in $V_t$. Next, our model computes $\phi(V_t)$ and $\psi(V_j)$ for all target segments in the video and updates the edges of the graph with the transition probabilities, given by sim$(\phi(V_t), \psi(V_j))$. 

Given that we fit the model on a single video, it is important that we ensure there is enough entropy in the transition distribution in order to ensure diversity in samples synthesized during inference. Always selecting the target segment with the highest transition probability would regurgitate the original sequence, as the model was trained to predict $V_{j+1}$ as the positive segment given $V_j$ as the query. Thus, given the current segment $V_j$, while we could transition to the very next segment $V_{j+1}$, we want to encourage the model to transition to other segments similar to $V_{j+1}$. 
While we assume that our input video sequence exhibits sufficient hierarchical, periodic structure to ensure repetition and multi-modality, we can also directly adjust the conditional entropy of the model through the softmax temperature term $\tau$. A lower temperature would flatten the transition probabilities (\ie increase the entropy) and reduce the difference in probabilities of the positive segment and segments similar to it. To avoid abrupt and noisy transitions, we set all transition probabilities below a certain threshold to zero. The threshold is set to be $t\%$ of the maximum transition probability connecting $V_j$ to any other node $V_t$. We compute $t$ heuristically and include details in Supp.

Next, we randomly select a positive segment to transition to from the edges with non-zero probabilities. This introduces variance in the generated textures and also ensures that the transitions are smooth and coherent. We then append the last $s$ number of frames in the positive segment to the output. This predicted positive segment $V_{t+1}$ is again fed into the network as the query and this is repeated to generate the whole output in an autoregressive fashion. 

\medskip
\noindent\textbf{Video Encoding.} We use the SlowFast~\citep{feichtenhofer2019slowfast} action recognition model pretrained on Kinetics-400~\citep{kay2017kinetics} for encoding the video segments. We include more details in Supp.

\medskip
\noindent\textbf{Interpolation.} For smoother transitions, we also conditionally interpolate between frames of the synthesized texture when there are transitions to different parts of the video. We use a pre-trained interpolation network of Jiang~\etal~\cite{jiang2018super}. In Sec.~\ref{sec:intp}, we include results both with and without interpolation to show that interpolation helps with smoothing. 

\section{Audio-Conditioned Video Textures} 
\label{sec:acvt}
We show that it is easy to extend our Contrastive Video Textures algorithm to synthesize videos that match a conditioning audio signal. Given an input video with corresponding audio $A^s$ and an external conditioning audio $A^c$, we synthesize a new video that is synchronized with the conditioning audio. We extract $N$ overlapping segments from the input and conditioning audio, as before. We compute the similarity of the input audio segments $A^s$ to the conditioning audio segment $A^c$ by projecting them into a common embedding space. We construct a transition probability matrix $T_a$ in the audio space as, 
$$T_a(i,j) = \mathrm{sim}(\varphi(A^c_i), \varphi(A^s_j))$$

Note that, unlike video segments in Eq.~\ref{eq:video_enc}, the audio segments come from two separate audio signals. Hence, there's no need to have two separate subnetworks as there's no symmetry and we use the same audio encoder $\varphi$ for both. We compute the transition probabilities $T_v$ for the target video segments given the previous predicted segment using the Contrastive video textures model (Eq.~\ref{eq:likelihood}). The joint transition probabilities for a segment are formulated as a trade-off between the audio-conditioning signal and the temporal coherence constraint as, 
\begin{equation}
    T = \alpha T_v + (1 - \alpha) T_a
    \label{eq:tradeoff}
\end{equation}

\noindent\textbf{Audio Encoding.}
We embed the audio segments using the VGGish model~\citep{hershey2017cnn} pretrained on AudioSet~\citep{gemmeke2017audio}. 

We describe implementation details of our method and hyperparameter choices in Supp.

\section{Experiments}
\label{sec:exp}
\begin{table}[t]
    \centering
    \begin{tabular}{lr@{\hskip2pt}c@{\hskip2pt}l}\toprule
        {\bf Method} & \multicolumn{3}{c}{\bf Preference \%}
        \\ \midrule 
        Classic & 3.33 & $\pm$ & 2.42 \ \% \\
        Classic Deep & 6.66 & $\pm$ & 3.37 \% \\
        Classic+ & 10.95 & $\pm$ & 4.22 \% \\
        Classic++ & 9.52 & $\pm$ & 3.97 \% \\ \cdashline{1-4}
        Any Classic & 30.48 & $\pm$ & 6.22 \% \\ \hline
        Contrastive & \textbf{69.52} & $\pm$ & 6.22 \% \\\bottomrule
    \end{tabular}
    \caption{\textbf{Perceptual Studies for Unconditional Video Textures}. We show MTurk evaluators textures synthesized by all 5 methods and ask them to pick the most realistic one. We also report the chance evaluators chose {\em any} of the variation of the classic model.}
    \label{tab:uncond}
\end{table}

\begin{table}[t]
    \centering
    \begin{tabular}{lr@{\hskip2pt}c@{\hskip2pt}l}\toprule
    {\bf Method}  & \multicolumn{3}{c}{\bf Real vs. Fake}
    \\ \midrule 
    Classic++ & 11.4 & $\pm$ & 4.30\%  \\
    Classic+ & 15.7 & $\pm$ & 4.92  \%\\\midrule
    Contrastive & \textbf{25.7} & $\pm$ & 4.30\% \\\bottomrule
    \end{tabular}
    \caption{\textbf{Unconditional: Real vs. Fake study.} We show evaluators a pair of videos (generated and real video) without labels, ask them to pick the real one. Our method fools evaluators more times than Classic.}
    \label{tab:uncond_rvf}
\end{table}

\begin{table}[t]
    \centering
    \begin{tabular}{lc}\toprule
    {\bf Method}  & {\bf Diversity Score}
    \\ \midrule 
    Classic+ & 2.30\\
    Classic++ & 2.66  \\\midrule
    Contrastive & \textbf{8.19} \\\bottomrule
    \end{tabular}
    \caption{\textbf{Unconditional: Diversity Scores.} We report the diversity scores obtained by our method and the two best baselines.}
    \label{tab:uncond_ds}
\end{table}

We curate a dataset of 70 videos from different domains such as dance and musical instruments including piano, guitar, suitar, tabla, flute, ukelele, and harmonium. A subset of these videos were randomly sampled from the PianoYT dataset~\citep{koepke2020sight} and the rest were downloaded from YouTube. We used 40 (of 70) videos to tune our hyperparameters and tested on the remaining 30 with no additional tuning. Our dataset consists of both short videos which are 2-3 minutes long and long videos ranging from 30-60 mins. We conduct perceptual evaluations on Amazon MTurk to qualitatively compare the results from our method to different baselines for both the unconditional and conditional settings. We also include results of ablating the interpolation module. Additionally, we introduce and report results on a new metric, diversity score, which measures the diversity of the textures.    

\begin{figure*}[t]
\centering
\begin{subfigure}{.3\textwidth}
    \centering
    \href{https://medhini.github.io/audio_video_textures/arxiv_videos/6.mp4}{\includegraphics[width=\linewidth]{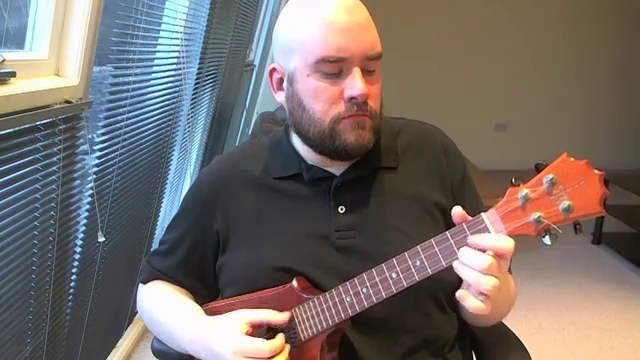}}
    \caption{Input Video}
    \label{fig:source}
\end{subfigure}
\begin{subfigure}{.3\textwidth}
    \centering
    \href{https://medhini.github.io/audio_video_textures/arxiv_videos/contrastive.mp4}{\includegraphics[width=\linewidth]{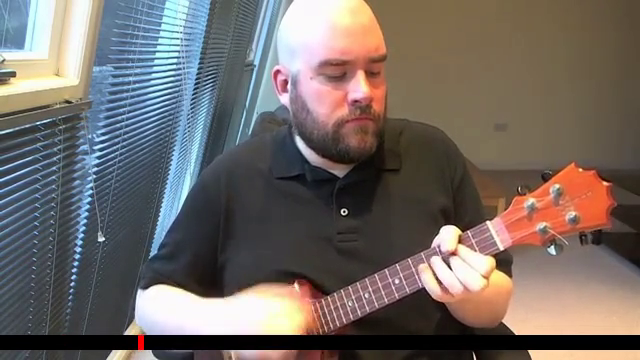}}
    \caption{Contrastive Video Texture}
    \label{fig:contrastive}
\end{subfigure}
\begin{subfigure}{.3\textwidth}
    \centering
    \href{https://medhini.github.io/audio_video_textures/arxiv_videos/classic.mp4}{\includegraphics[width=\linewidth]{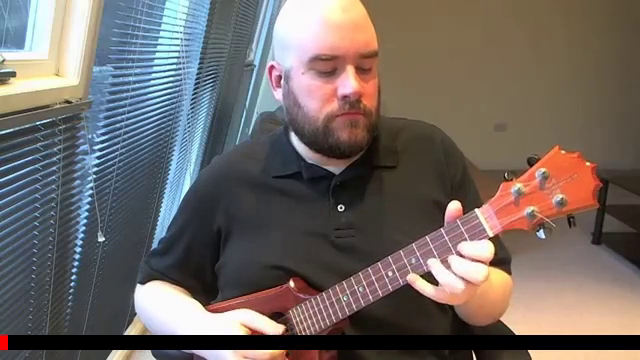}}
    \caption{Classic Video Texture}
    \label{fig:classic}
\end{subfigure}
\medskip
\begin{subfigure}{.3\textwidth}
    \centering
    \href{https://medhini.github.io/audio_video_textures/arxiv_videos/classicplus.mp4}{\includegraphics[width=\linewidth]{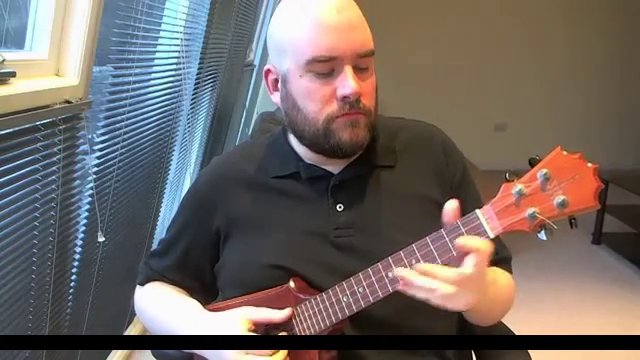}}
    \caption{Classic+ Video Texture}
    \label{fig:classic+}
\end{subfigure}
\begin{subfigure}{.3\textwidth}
    \centering
    \href{https://medhini.github.io/audio_video_textures/arxiv_videos/classicplusplus.mp4}{\includegraphics[width=\linewidth]{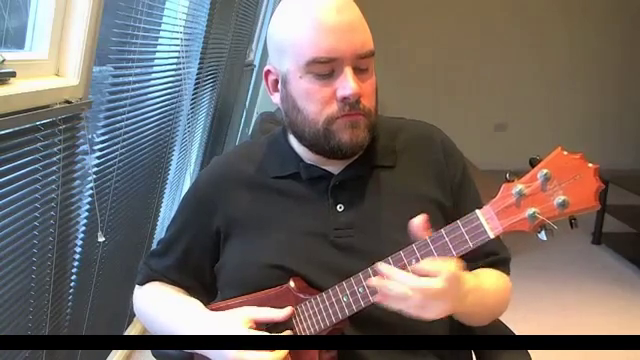}}
    \caption{Classic++ Video Texture}
    \label{fig:classic++}
\end{subfigure}
\begin{subfigure}{.3\textwidth}
    \centering
    \href{https://medhini.github.io/audio_video_textures/arxiv_videos/classic_deep.mp4}{\includegraphics[width=\linewidth]{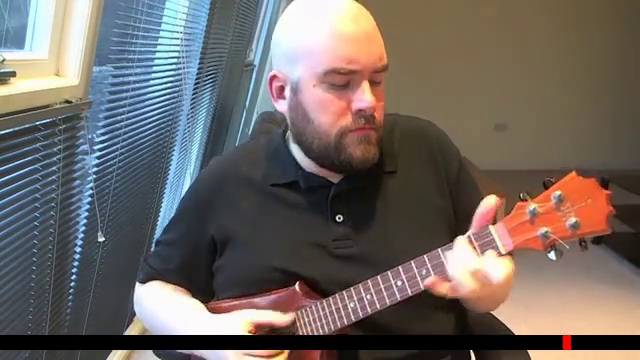}}
    \caption{Classic Deep Video Texture}
    \label{fig:classic_deep}
\end{subfigure}
\caption{\textbf{Unconditional Contrastive Video Textures.} Click on each figure to play the video. The figure shows the input video and textures synthesized using our Contrastive method and the baselines Classic, Classic+, Classic++, and Classic Deep. The red bar at the bottom of each video indicates the part of the input video being played. Classic, Classic+, and Classic++ textures loop over the same frame at the start of the video as shown by the red bar, are choppy, and not diverse. Classic Deep texture has jarring transitions. Our Contrastive method finds smooth and seamless transitions in the video to produce a texture that's diverse yet dynamically consistent.}
\label{fig:uncond}
\end{figure*}

\subsection{Unconditional Video Texture Synthesis}
\label{sec:intp}
To show the effectiveness of our method, we compare our results to the Classic video textures algorithm~\citep{schodl2000video} and its three variations. The algorithm and its variants are described in Sec. 1 of Supplementary. Classic+, like Contrastive, appends multiple frames to the output sequence instead of a single frame, Classic++ adds a stride while filtering the distance matrix and Classic Deep uses ImageNet pretrained ResNet features instead of raw pixel values. For fairness, we added the interpolation module described in Sec.~\ref{sec:cvt} to all the baselines. 

Table~\ref{tab:uncond} reports the results from a perceptual study on Amazon MTurk where evaluators were shown textures generated by all five methods and asked to choose the one they found most realistic. Our Contrastive model surpasses all baselines by a large margin and was chosen 69.52\% of the time. Since the classic models are similar, we also report all variations of classic combined. They are chosen 30.48\% of the time. 

\begin{figure}[t]
    \centering
    \begin{adjustbox}{max width=\linewidth}
    \setlength{\tabcolsep}{1pt}
    \begin{tabular}{cc}
         \href{https://medhini.github.io/audio_video_textures/arxiv_videos/guitar1.mp4}{\includegraphics[width=0.5\linewidth]{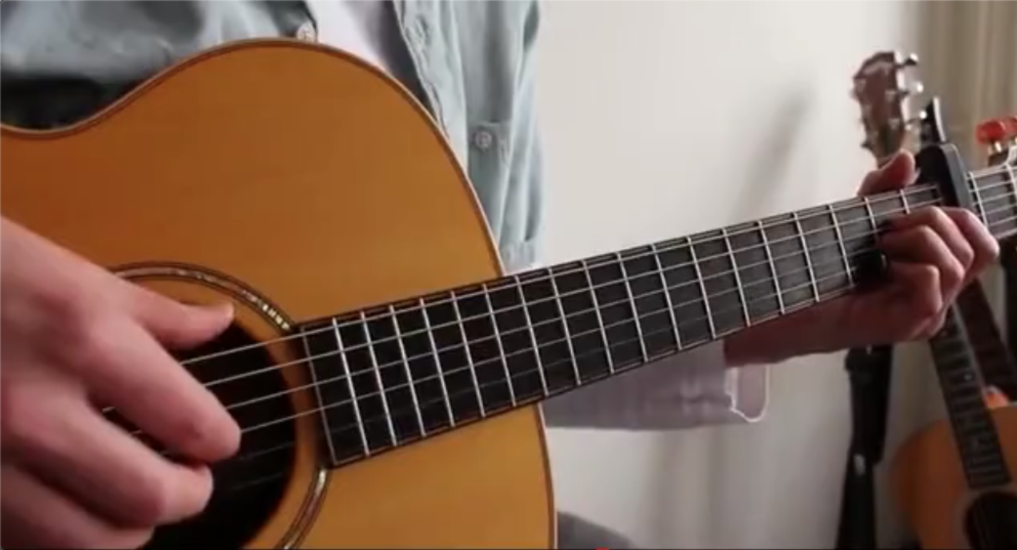}} & 
         \href{https://medhini.github.io/audio_video_textures/arxiv_videos/pinkpanther.mp4}{\includegraphics[width=0.5\linewidth]{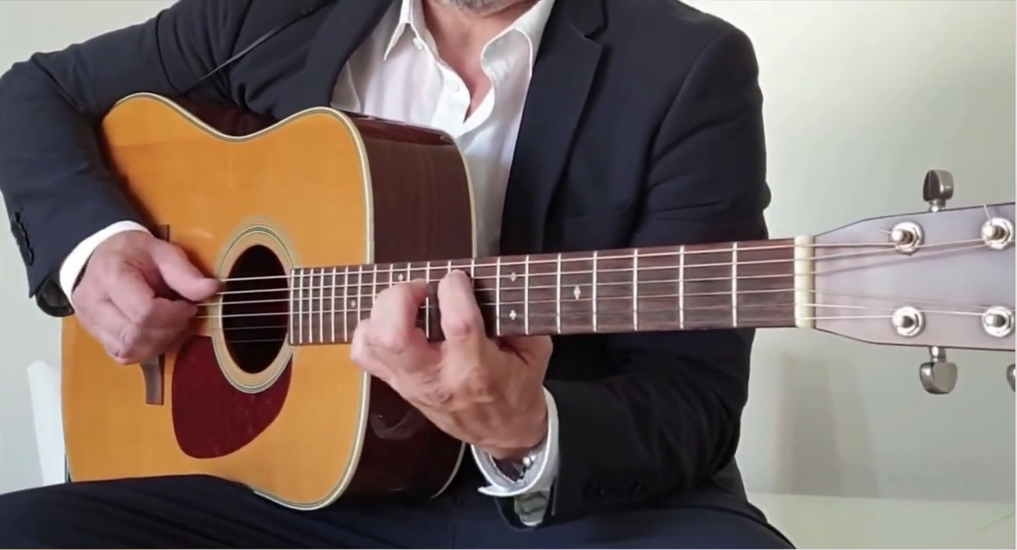}}\\
         \multicolumn{2}{c}{(a) Guitar} \\
         \href{https://medhini.github.io/audio_video_textures/arxiv_videos/afro.mp4}{\includegraphics[width=0.5\linewidth,cfbox=black 0.5pt 0pt]{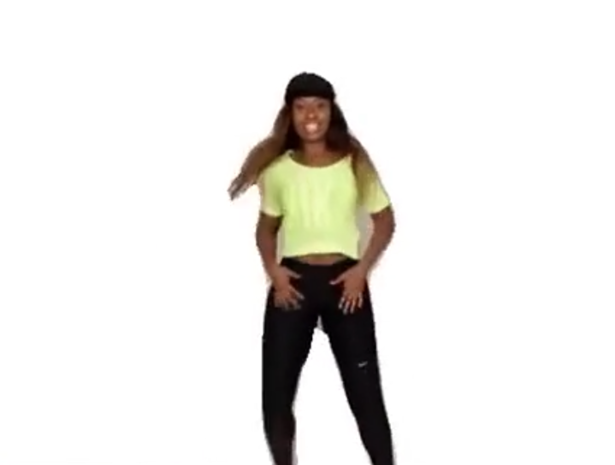}} & \href{https://medhini.github.io/audio_video_textures/arxiv_videos/lisa2.mp4}{\includegraphics[width=0.5\linewidth]{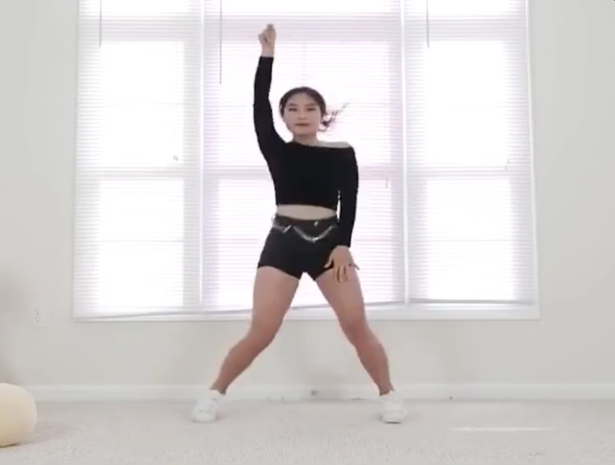}} \\
         \multicolumn{2}{c}{(b) Dance} \\
         \href{https://medhini.github.io/audio_video_textures/arxiv_videos/tabla.mp4}{\includegraphics[width=0.5\linewidth]{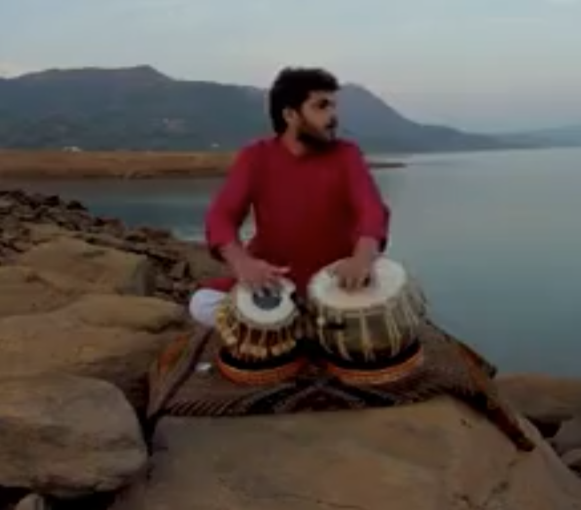}} & \href{https://medhini.github.io/audio_video_textures/arxiv_videos/sitar.mp4}{\includegraphics[width=0.5\linewidth]{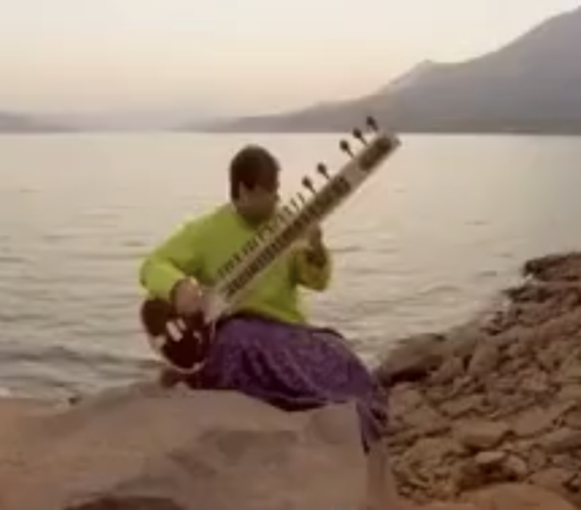}} \\
         \multicolumn{2}{c}{(c) Indian Musical Instruments (Tabla and Sitar)} \\
         \multicolumn{2}{c}{\href{https://medhini.github.io/audio_video_textures/arxiv_videos/harp.mp4}{\includegraphics[width=0.7\linewidth]{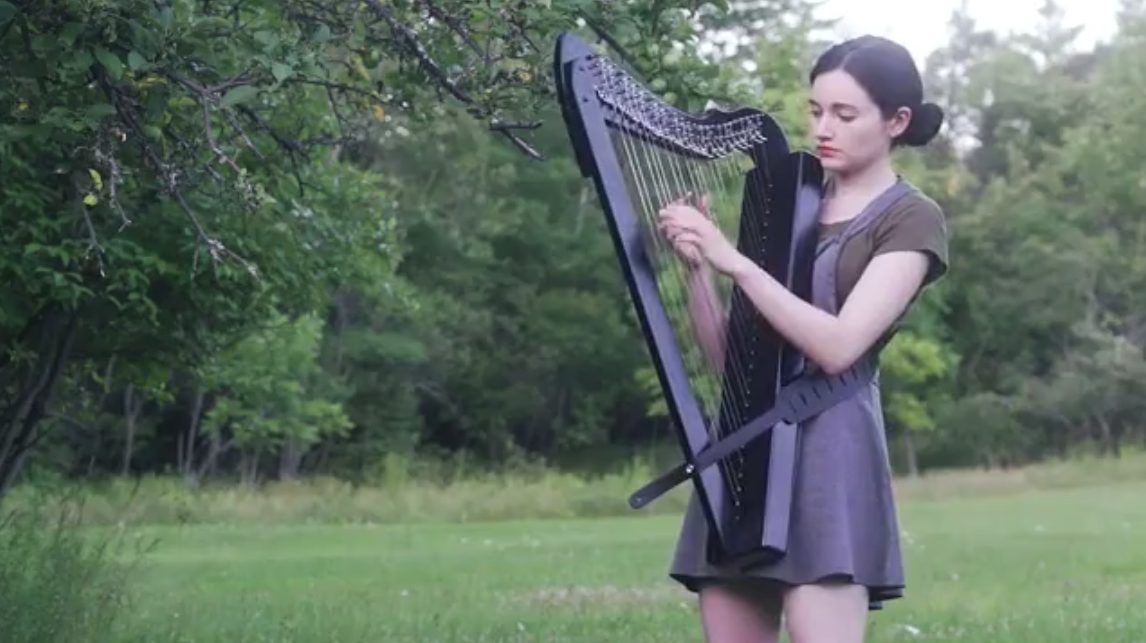}}}\\
         \multicolumn{2}{c}{(d) Harp} \\
    \end{tabular}
    \end{adjustbox}
    \captionof{figure}{Qualitative Results of \textbf{Unconditional Contrastive Video Textures}. Click on the image to play the video.}
    \label{fig:contres}
\end{figure}

We include qualitative video results for Contrastive, Classic, Classic+, Classic++, and Classic Deep in Fig.~\ref{fig:uncond}. The red bar at the bottom of each video indicates the part of the input video being played. Video in Fig.~\ref{fig:contrastive} produced by our Contrastive method is the most realistic and consistent with seamless transitions. The red bar transitioning to new positions indicates that our textures are dynamic and diverse (vary over time). Classic in Fig.~\ref{fig:classic} loops over the same set of frames in and around the target, thus appearing stuck. Classic+ and Classic++ shown in Fig.~\ref{fig:classic+} and Fig.~\ref{fig:classic++} have slightly improved quality compared to it but lack diversity and produce jarring transitions. Classic Deep texture in Fig.~\ref{fig:classic_deep} is choppy due to multiple poor transitions chosen by the model. 

\begin{table}[t]
    \centering
    \begin{tabular}{lr@{\hskip2pt}c@{\hskip2pt}l}\toprule
    {\bf Method}  &\multicolumn{3}{c}{\bf Real vs Fake}
    \\ \midrule 
    Random Clip & 15.33 & $\pm$ & 5.76\% \\
    Audio NN & 20.4 & $\pm$ & 6.63\% \\\midrule
    Contrastive & \textbf{26.74} & $\pm$ & 6.14\%\\\bottomrule
    \end{tabular}
    \caption{\textbf{Conditional: Real vs. Fake study.} We show evaluators a pair of videos (generated and real video) without labels and ask them to pick the real one. Our method fooled evaluators more often than the baselines.}
    \label{tab:cond}
\end{table}

Additionally, we conduct real vs. fake studies in Table~\ref{tab:uncond_rvf} where the evaluators are shown the ground truth video and synthesized texture and asked to pick the one they think is real. Our method is able to fool evaluators 25.7\% of the time whereas the best baseline (Classic+) is able to fool the evaluators only 15.7\% of the time.

Both the qualitative and quantitative comparisons clearly highlight the issues with the Classic model and emphasize the need to learn the feature representations and the distance metric as we do in our Contrastive method. Fig.~\ref{fig:contres} shows more qualitative results of Unconditional Contrastive Video Textures on videos of guitar, dance, Indian musical instruments, and the harp. Our method works well on all the domains and produces dynamic yet consistent video textures. The change in position of the red bar indicates that our method seamlessly transitions across different parts of the input video. As seen in the dance videos, learned representations result in transitions that are consistent with the arm movements of the dancer. 

\noindent\textbf{Diversity Score.} For a fair comparison, we set the temperature for Contrastive and Classic+ methods such that the resulting videos have approximately the same number of transitions. To do this, we grid search over a range of temperatures and count the transitions produced by each. We synthesize Classic+ and Contrastive video textures with 10$\pm$2 transitions each for 20 videos. Evaluators were shown textures from both methods and asked to pick the one they found more realistic. Contrastive videos were preferred 76.6\% of the time, comparable to the result in Tab.~\ref{tab:uncond}, indicating that our method finds better transitions.

Keeping the number of transitions fixed (10$\pm$2), we measure diversity score (DS) as the number of new transitions in every 30 seconds of the synthesized video, averaged over all videos. A transition is considered \emph{new} if it hasn't occurred in the 30 second time-frame. Tab.~\ref{tab:uncond_ds} shows the scores obtained by our method and the two best baselines (Classic+ and Classic++). Our method achieves a DS of 8.19, indicating that our textures are diverse and contain, on an average, 8 (of 10) new transitions every 30 seconds. Classic+ and Classic++ achieve diversity scores 2.30 and and 2.66 respectively, indicating that for both methods the video texture loops over the same part of the input video.

\noindent\textbf{Interpolation.} 
We verify the effectiveness of the interpolation module through a perceptual study. Evaluators were shown two videos (with and without interpolation), and asked to pick the one they found more realistic. They picked the video with interpolation 89\% of the time, thus confirming that interpolation leads to an improvement in perceptual quality. We show qualitative comparisons \href{https://sites.google.com/view/extracontrastivevt2021/home?authuser=2#h.yt2qctwyutke}{\underline{here}}).

\subsection{Audio-Conditioned Contrastive Video Textures}

For audio-conditioned video synthesis, we randomly paired the 70 videos with songs from the same domain (\emph{e.g.} a input piano video is paired with a conditioning audio of a piano). Using this strategy, we created 70 input video - conditioning audio pairs. As described in Sec.~\ref{sec:acvt}, we extend our Contrastive method to synthesize textures given a conditioning audio signal. We compare audio-conditioned video textures synthesized by our method to four baselines and report results from a perceptual evaluation. 

\smallskip \noindent $\sbullet$ \emph{Random Clip.} In this baseline, we choose a random portion of the input video to match the conditioning audio. 

\smallskip \noindent $\sbullet$ \emph{Classic+Audio.} We add audio-conditioning to the classic video textures algorithm. For this, we divide the conditioning audio into segments and find nearest-neighbours in the input audio. Then we combine these distances with the distance matrix calculated by video textures using Eq.~\ref{eq:tradeoff}. 

\smallskip \noindent $\sbullet$ \emph{Visual Rhythm and Beat(VRB).} We use the approach of Davis~\etal~\cite{davis2018visual} to synchronize the input video with the audio beats. This method works by changing the pace of the video to better align the visual and audio beats.

\smallskip \noindent $\sbullet$ \emph{Audio Nearest Neighbours.} We include comparisons to a nearest neighbor baseline that works by computing the similarity between the conditioning audio signal and segments of the input audio of the same length, and then choosing the video clip of the closest match.

\begin{figure}[t]
    \centering
    \includegraphics[width=0.4\textwidth]{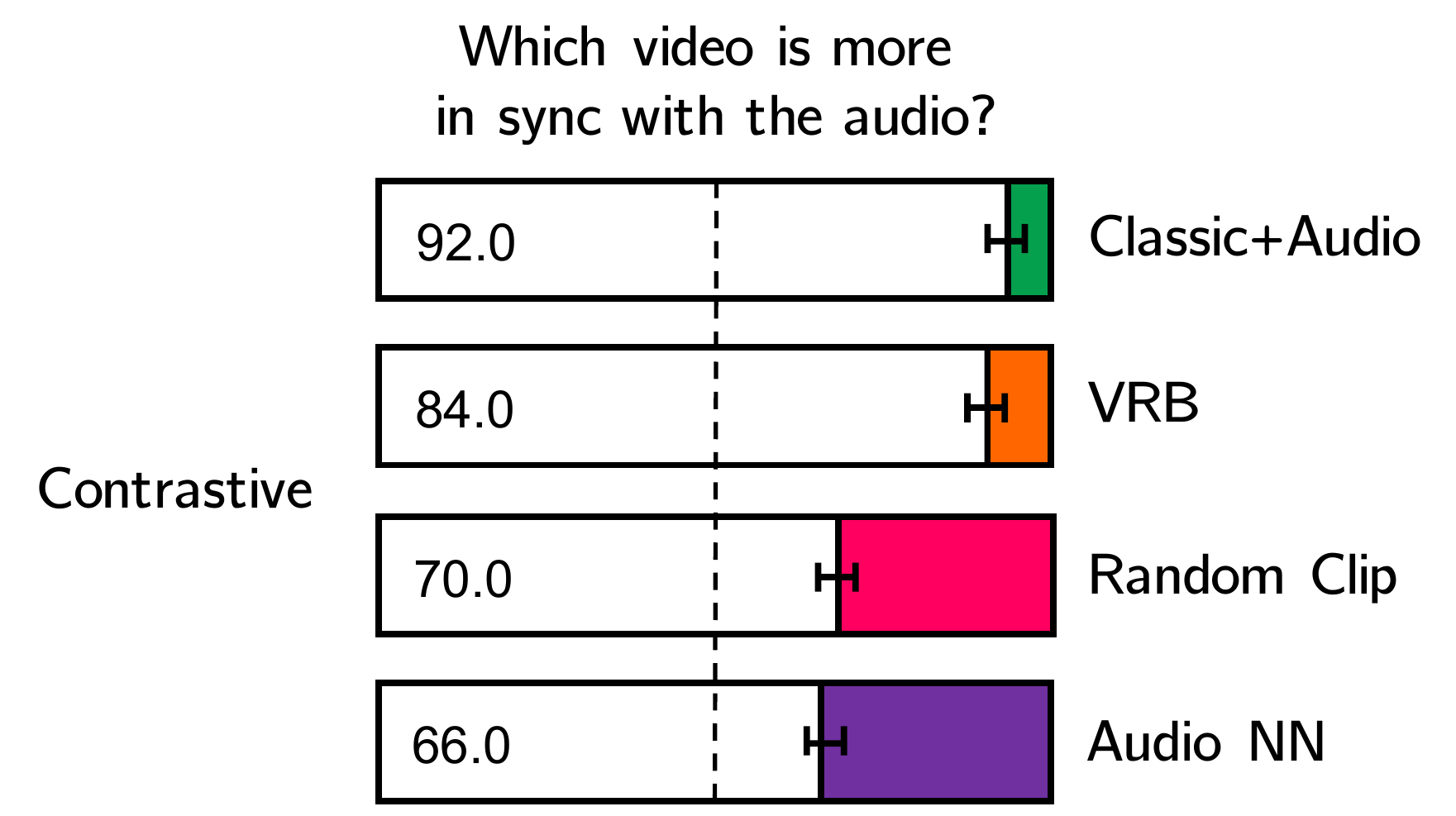}
    \caption{\textbf{Perceptual Studies for Audio-Conditioned Contrastive Video Textures.} We compare results from our Contrastive method against each of the baselines, individually. Evaluators were shown two videos, one synthesized by our method and the other by the corresponding baseline and asked to pick the one where the audio and video were more in sync.}
    \label{fig:cond}
\end{figure}

\begin{figure*}[h]
    \centering
    \includegraphics[width=\textwidth]{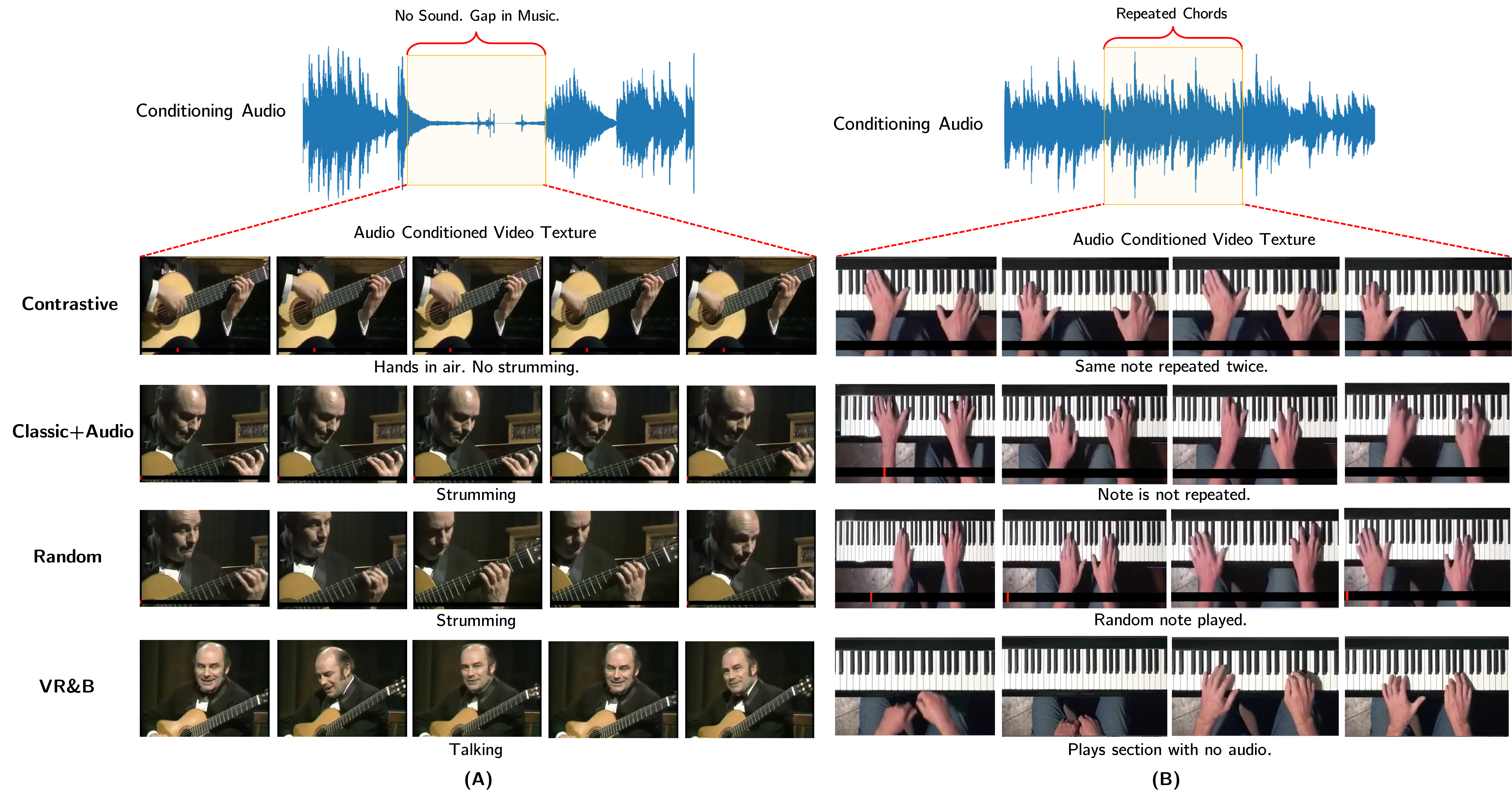}
    \caption{Qualitative comparison of audio-conditioned video textures synthesized by Classic+Audio, Random Clip, Visual Rhythm and Beat (VRB) and our Contrastive model. \textbf{(A)} The conditioning audio waveform shows a gap in the audio where no music is being played. Our model is able to pick up on that and the corresponding video that is synthesized has hands in the air and no strumming. However, both Random Clip and Classic+Audio show strumming, and VRB shows the person talking. \textbf{(B)} The conditioning audio waveform has the same chord repeated twice. The video synthesized by our model reflects this, and we observe the same frames (1 and 2) repeated again. Classic+Audio and Random Clip don't repeat the note and VRB result contains a region without audio where the person isn't playing anything.}
    \label{fig:audio_gap}
\end{figure*}

Fig.~\ref{fig:cond} shows results from our perceptual studies comparing the audio-conditioned video textures synthesized by our Contrastive model to all of the baselines. The evaluators were shown two videos with the same conditioning audio, one synthesized by our method and the other by the corresponding baseline. They were asked to pick the video that was more in sync with the audio. Our method outperforms all baselines by a large margin. As shown in Tab.~\ref{tab:cond}, we conducted a real vs. fake study comparing the ground truth videos with the synthesized videos from Contrastive and the two best baselines (Random Clip and Audio NN). While Random Clip and Audio NN beat the ground truth only 15.33\% and 20.4\% respectively, our method was able to fool evaluators 26.74\% of the time. 

Fig.~\ref{fig:audio_gap} shows qualitative results comparing our method to the five baselines described above. We are also include the videos \href{https://sites.google.com/view/contrastivevt2021/home?authuser=2#h.zf4usrxjx4hz}{\underline{here}}. In Fig.~\ref{fig:audio_gap} (A), the conditioning audio signal has a gap/break in the audio where no music is being played. We see the output produced by our Contrastive model is semantically meaningful and aligns best with the audio. Random Clip chooses a random segment which has strumming and thus fails to align with the audio. Similarly Classic+Audio chooses frames that don't correlate with the audio. VRB doesn't capture semantics as it only speeds up or slows down the video to better match the audio beats. Similarly, in Fig.~\ref{fig:audio_gap}(B) we see that our Contrastive method is able to pick up on repeated chords in the conditioning audio signal while no other method is able to do that. Through more examples listed at this \href{https://sites.google.com/view/contrastivevt2021/home?authuser=2#h.dn47d0tdrtle}{\underline{website}}, we show that the videos synthesized by Contrastive model are more in sync with the conditioning audio. For example, it identifies gaps in the audio, repeated chords, and change of pace. We observed experimentally that our method doesn't work well for videos where the scene constantly changes (such as waves) and where subtle asynchronies between audio and video are easy to spot (such as people speaking) as these applications are beyond the scope of video textures. We hope that they can be addressed with hybrid approaches that combine the benefits of video textures with GANs.

\subsection{Comparison to Video Generation Methods} GAN based video generation methods cannot capture the temporal dynamics of the video and thus fail to synthesize long and diverse textures. We compare our unconditional Contrastive method to MoCoGAN~\citep{tulyakov2018mocogan}, an unconditional video generation method and include results \href{https://sites.google.com/view/extracontrastivevt2021/home?authuser=2#h.8e0dxjgknl8u}{\underline{here}}. A 3-second video of a candle flame is given as input and our method is able to produce a 30-second high resolution, temporally consistent, and diverse output of a candle flickering. On the other hand, MoCoGAN's video contains artifacts and the flame lasts for only 3 seconds. Similarly, with the guitar video, our method produces a realistic video with seamless transitions whereas MoCoGAN's output is blurry.

\section{Conclusion}

We presented Contrastive Video Textures, a learning-based approach for video textures applied to audio-conditioned video synthesis. Our method fits an input-specific bi-gram model to capture the dynamics of a video, and uses it to generate diverse and temporally coherent textures. We also introduced audio-conditioned video texture synthesis as a useful application of video textures. We show that our model outperforms a number of baselines on perceptual studies. 

\medskip\noindent\textbf{Acknowledgements.} We thank Arun Mallya, Allan Jabri, Anna Rohrbach, Amir Bar, Suzie Petryk, and Parsa Mahmoudieh for very helpful discussions and feedback. This work was supported in part by DoD including DARPA's XAI, LwLL, and SemaFor programs, as well as BAIR's industrial alliance programs.
{\small
\bibliographystyle{ieee_fullname}
\bibliography{egbib}

\begin{thebibliography}{10}\itemsep=-1pt

\bibitem{chen2020simple}
Ting Chen, Simon Kornblith, Mohammad Norouzi, and Geoffrey Hinton.
\newblock A simple framework for contrastive learning of visual
  representations.
\newblock {\em arXiv preprint arXiv:2002.05709}, 2020.

\bibitem{chen2020improved}
Xinlei Chen, Haoqi Fan, Ross Girshick, and Kaiming He.
\newblock Improved baselines with momentum contrastive learning.
\newblock {\em arXiv preprint arXiv:2003.04297}, 2020.

\bibitem{chen2019mocycle}
Yang Chen, Yingwei Pan, Ting Yao, Xinmei Tian, and Tao Mei.
\newblock Mocycle-gan: Unpaired video-to-video translation.
\newblock 2019.

\bibitem{clark2019efficient}
Aidan Clark, Jeff Donahue, and Karen Simonyan.
\newblock Efficient video generation on complex datasets.
\newblock {\em arXiv preprint arXiv:1907.06571}, 2019.

\bibitem{davis2018visual}
Abe Davis and Maneesh Agrawala.
\newblock Visual rhythm and beat.
\newblock {\em ACM Transactions on Graphics (TOG)}, 2018.

\bibitem{denton2017unsupervised}
Emily~L Denton and Vighnesh Birodkar.
\newblock Unsupervised learning of disentangled representations from video.
\newblock In {\em Advances in Neural Information Processing Systems (NeurIPS)},
  2017.

\bibitem{efros2003}
Alexei~A. Efros, Alexander~C. Berg, Greg Mori, and Jitendra Malik.
\newblock Recognizing action at a distance.
\newblock In {\em IEEE International Conference on Computer Vision (ICCV)},
  pages 726--733, Nice, France, 2003.

\bibitem{efros2001image}
Alexei~A Efros and William~T Freeman.
\newblock Image quilting for texture synthesis and transfer.
\newblock In {\em ACM SIGGRAPH}, 2001.

\bibitem{efros1999texture}
Alexei~A Efros and Thomas~K Leung.
\newblock Texture synthesis by non-parametric sampling.
\newblock In {\em IEEE International Conference on Computer Vision (ICCV)},
  1999.

\bibitem{ephrat2018looking}
Ariel Ephrat, Inbar Mosseri, Oran Lang, Tali Dekel, Kevin Wilson, Avinatan
  Hassidim, William~T Freeman, and Michael Rubinstein.
\newblock Looking to listen at the cocktail party: A speaker-independent
  audio-visual model for speech separation.
\newblock {\em ACM SIGGRAPH}, 2018.

\bibitem{feichtenhofer2019slowfast}
Christoph Feichtenhofer, Haoqi Fan, Jitendra Malik, and Kaiming He.
\newblock Slowfast networks for video recognition.
\newblock In {\em IEEE International Conference on Computer Vision (ICCV)},
  2019.

\bibitem{gafni2019vid2game}
Oran Gafni, Lior Wolf, and Yaniv Taigman.
\newblock Vid2game: Controllable characters extracted from real-world videos.
\newblock In {\em International Conference on Learning Representations (ICLR)},
  2020.

\bibitem{gatys2015texture}
Leon Gatys, Alexander~S Ecker, and Matthias Bethge.
\newblock Texture synthesis using convolutional neural networks.
\newblock In {\em Advances in Neural Information Processing Systems (NeurIPS)},
  2015.

\bibitem{gemmeke2017audio}
Jort~F Gemmeke, Daniel~PW Ellis, Dylan Freedman, Aren Jansen, Wade Lawrence,
  R~Channing Moore, Manoj Plakal, and Marvin Ritter.
\newblock Audio set: An ontology and human-labeled dataset for audio events.
\newblock In {\em IEEE International Conference on Acoustics, Speech and Signal
  Processing (ICASSP)}. IEEE, 2017.

\bibitem{goodfellow2014generative}
Ian Goodfellow, Jean Pouget-Abadie, Mehdi Mirza, Bing Xu, David Warde-Farley,
  Sherjil Ozair, Aaron Courville, and Yoshua Bengio.
\newblock Generative adversarial networks.
\newblock In {\em Advances in Neural Information Processing Systems (NeurIPS)},
  2014.

\bibitem{gur2020hierarchical}
Shir Gur, Sagie Benaim, and Lior Wolf.
\newblock Hierarchical patch vae-gan: Generating diverse videos from a single
  sample.
\newblock {\em arXiv preprint arXiv:2006.12226}, 2020.

\bibitem{he2020momentum}
Kaiming He, Haoqi Fan, Yuxin Wu, Saining Xie, and Ross Girshick.
\newblock Momentum contrast for unsupervised visual representation learning.
\newblock In {\em IEEE Conference on Computer Vision and Pattern Recognition
  (CVPR)}, pages 9729--9738, 2020.

\bibitem{heeger1995pyramid}
David~J Heeger and James~R Bergen.
\newblock Pyramid-based texture analysis/synthesis.
\newblock In {\em ACM SIGGRAPH}, pages 229--238, 1995.

\bibitem{henaff2019data}
Olivier~J H{\'e}naff, Aravind Srinivas, Jeffrey De~Fauw, Ali Razavi, Carl
  Doersch, SM Eslami, and Aaron van~den Oord.
\newblock Data-efficient image recognition with contrastive predictive coding.
\newblock {\em arXiv preprint arXiv:1905.09272}, 2019.

\bibitem{hershey2017cnn}
Shawn Hershey, Sourish Chaudhuri, Daniel~PW Ellis, Jort~F Gemmeke, Aren Jansen,
  R~Channing Moore, Manoj Plakal, Devin Platt, Rif~A Saurous, Bryan Seybold,
  et~al.
\newblock Cnn architectures for large-scale audio classification.
\newblock In {\em IEEE International Conference on Acoustics, Speech and Signal
  Processing (ICASSP)}, 2017.

\bibitem{holynski2020animating}
Aleksander Holynski, Brian Curless, Steven~M Seitz, and Richard Szeliski.
\newblock Animating pictures with eulerian motion fields.
\newblock {\em arXiv preprint arXiv:2011.15128}, 2020.

\bibitem{jiang2018super}
Huaizu Jiang, Deqing Sun, Varun Jampani, Ming-Hsuan Yang, Erik Learned-Miller,
  and Jan Kautz.
\newblock Super slomo: High quality estimation of multiple intermediate frames
  for video interpolation.
\newblock In {\em IEEE Conference on Computer Vision and Pattern Recognition
  (CVPR)}, 2018.

\bibitem{kalchbrenner2016video}
Nal Kalchbrenner, Aaron van~den Oord, Karen Simonyan, Ivo Danihelka, Oriol
  Vinyals, Alex Graves, and Koray Kavukcuoglu.
\newblock Video pixel networks.
\newblock {\em International Conference on Machine Learning (ICML)}, 2017.

\bibitem{karras2018style}
Tero Karras, Samuli Laine, and Timo Aila.
\newblock A style-based generator architecture for generative adversarial
  networks.
\newblock In {\em IEEE Conference on Computer Vision and Pattern Recognition
  (CVPR)}, 2019.

\bibitem{kay2017kinetics}
Will Kay, Joao Carreira, Karen Simonyan, Brian Zhang, Chloe Hillier, Sudheendra
  Vijayanarasimhan, Fabio Viola, Tim Green, Trevor Back, Paul Natsev, et~al.
\newblock The kinetics human action video dataset.
\newblock {\em arXiv preprint arXiv:1705.06950}, 2017.

\bibitem{kim2018learning}
Changil Kim, Hijung~Valentina Shin, Tae-Hyun Oh, Alexandre Kaspar, Mohamed
  Elgharib, and Wojciech Matusik.
\newblock On learning associations of faces and voices.
\newblock In {\em Asian Conference on Computer Vision}, 2018.

\bibitem{kingma2013auto}
Diederik~P Kingma and Max Welling.
\newblock Auto-encoding variational {Bayes}.
\newblock In {\em International Conference on Learning Representations (ICLR)},
  2013.

\bibitem{koepke2020sight}
A~Sophia Koepke, Olivia Wiles, Yael Moses, and Andrew Zisserman.
\newblock Sight to sound: An end-to-end approach for visual piano
  transcription.
\newblock In {\em IEEE International Conference on Acoustics, Speech and Signal
  Processing (ICASSP)}. IEEE, 2020.

\bibitem{kwatra2003graphcut}
Vivek Kwatra, Arno Sch{\"o}dl, Irfan Essa, Greg Turk, and Aaron Bobick.
\newblock Graphcut textures: image and video synthesis using graph cuts.
\newblock {\em ACM Transactions on Graphics (TOG)}, 2003.

\bibitem{lee2019dancing}
Hsin-Ying Lee, Xiaodong Yang, Ming-Yu Liu, Ting-Chun Wang, Yu-Ding Lu,
  Ming-Hsuan Yang, and Jan Kautz.
\newblock Dancing to music.
\newblock In {\em Advances in Neural Information Processing Systems (NeurIPS)},
  2019.

\bibitem{li2018video}
Yitong Li, Martin~Renqiang Min, Dinghan Shen, David~E Carlson, and Lawrence
  Carin.
\newblock Video generation from text.
\newblock In {\em AAAI}, 2018.

\bibitem{mallya2020world}
Arun Mallya, Ting-Chun Wang, Karan Sapra, and Ming-Yu Liu.
\newblock World-consistent video-to-video synthesis.
\newblock In {\em European Conference on Computer Vision (ECCV)}, 2020.

\bibitem{menapace2021playable}
Willi Menapace, St{\'e}phane Lathuili{\`e}re, Sergey Tulyakov, Aliaksandr
  Siarohin, and Elisa Ricci.
\newblock Playable video generation.
\newblock {\em arXiv preprint arXiv:2101.12195}, 2021.

\bibitem{mikolov2013distributed}
Tomas Mikolov, Ilya Sutskever, Kai Chen, Greg~S Corrado, and Jeff Dean.
\newblock Distributed representations of words and phrases and their
  compositionality.
\newblock In {\em Advances in Neural Information Processing Systems (NeurIPS)},
  2013.

\bibitem{misra2016shuffle}
Ishan Misra, C~Lawrence Zitnick, and Martial Hebert.
\newblock Shuffle and learn: unsupervised learning using temporal order
  verification.
\newblock In {\em European Conference on Computer Vision (ECCV)}. Springer,
  2016.

\bibitem{oh2019speech2face}
Tae-Hyun Oh, Tali Dekel, Changil Kim, Inbar Mosseri, William~T Freeman, Michael
  Rubinstein, and Wojciech Matusik.
\newblock Speech2face: Learning the face behind a voice.
\newblock In {\em IEEE Conference on Computer Vision and Pattern Recognition
  (CVPR)}, 2019.

\bibitem{oord2018representation}
Aaron van~den Oord, Yazhe Li, and Oriol Vinyals.
\newblock Representation learning with contrastive predictive coding.
\newblock {\em arXiv preprint arXiv:1807.03748}, 2018.

\bibitem{park2019semantic}
Taesung Park, Ming-Yu Liu, Ting-Chun Wang, and Jun-Yan Zhu.
\newblock Semantic image synthesis with spatially-adaptive normalization.
\newblock In {\em IEEE Conference on Computer Vision and Pattern Recognition
  (CVPR)}, 2019.

\bibitem{portilla2000parametric}
Javier Portilla and Eero~P Simoncelli.
\newblock A parametric texture model based on joint statistics of complex
  wavelet coefficients.
\newblock {\em International Journal of Computer Vision (IJCV)}, 2000.

\bibitem{saito2017temporal}
Masaki Saito, Eiichi Matsumoto, and Shunta Saito.
\newblock Temporal generative adversarial nets with singular value clipping.
\newblock In {\em IEEE International Conference on Computer Vision (ICCV)},
  2017.

\bibitem{schodl2001machine}
Arno Sch{\"o}dl and Irfan~A Essa.
\newblock Machine learning for video-based rendering.
\newblock In {\em Advances in Neural Information Processing Systems (NeurIPS)},
  2001.

\bibitem{schodl2002controlled}
Arno Sch{\"o}dl and Irfan~A Essa.
\newblock Controlled animation of video sprites.
\newblock In {\em ACM SIGGRAPH}, 2002.

\bibitem{schodl2000video}
Arno Sch{\"o}dl, Richard Szeliski, David~H Salesin, and Irfan Essa.
\newblock Video textures.
\newblock In {\em ACM SIGGRAPH}, 2000.

\bibitem{shaham2019singan}
Tamar~Rott Shaham, Tali Dekel, and Tomer Michaeli.
\newblock Singan: Learning a generative model from a single natural image.
\newblock In {\em IEEE International Conference on Computer Vision (ICCV)},
  2019.

\bibitem{srivastava2015unsupervised}
Nitish Srivastava, Elman Mansimov, and Ruslan Salakhudinov.
\newblock Unsupervised learning of video representations using {LSTMs}.
\newblock In {\em International Conference on Machine Learning (ICML)}, 2015.

\bibitem{tulyakov2018mocogan}
Sergey Tulyakov, Ming-Yu Liu, Xiaodong Yang, and Jan Kautz.
\newblock {MoCoGAN}: Decomposing motion and content for video generation.
\newblock In {\em IEEE Conference on Computer Vision and Pattern Recognition
  (CVPR)}, 2018.

\bibitem{ulyanov2018deep}
Dmitry Ulyanov, Andrea Vedaldi, and Victor Lempitsky.
\newblock Deep image prior.
\newblock In {\em IEEE Conference on Computer Vision and Pattern Recognition
  (CVPR)}, 2018.

\bibitem{unterthiner2018towards}
Thomas Unterthiner, Sjoerd van Steenkiste, Karol Kurach, Raphael Marinier,
  Marcin Michalski, and Sylvain Gelly.
\newblock Towards accurate generative models of video: A new metric \&
  challenges.
\newblock {\em arXiv preprint arXiv:1812.01717}, 2018.

\bibitem{vondrick2016generating}
Carl Vondrick, Hamed Pirsiavash, and Antonio Torralba.
\newblock Generating videos with scene dynamics.
\newblock In {\em Advances in Neural Information Processing Systems (NeurIPS)},
  2016.

\bibitem{wang2019few}
Ting-Chun Wang, Ming-Yu Liu, Andrew Tao, Guilin Liu, Jan Kautz, and Bryan
  Catanzaro.
\newblock Few-shot video-to-video synthesis.
\newblock In {\em Advances in Neural Information Processing Systems (NeurIPS)},
  2019.

\bibitem{wang2018vid2vid}
Ting-Chun Wang, Ming-Yu Liu, Jun-Yan Zhu, Guilin Liu, Andrew Tao, Jan Kautz,
  and Bryan Catanzaro.
\newblock Video-to-video synthesis.
\newblock In {\em Advances in Neural Information Processing Systems (NeurIPS)},
  2018.

\bibitem{wang2018video}
Ting-Chun Wang, Ming-Yu Liu, Jun-Yan Zhu, Guilin Liu, Andrew Tao, Jan Kautz,
  and Bryan Catanzaro.
\newblock Video-to-video synthesis.
\newblock In {\em Advances in Neural Information Processing Systems (NeurIPS)},
  2018.

\bibitem{wei2018learning}
Donglai Wei, Joseph~J Lim, Andrew Zisserman, and William~T Freeman.
\newblock Learning and using the arrow of time.
\newblock In {\em IEEE Conference on Computer Vision and Pattern Recognition
  (CVPR)}, pages 8052--8060, 2018.

\bibitem{wei2009state}
Li-Yi Wei, Sylvain Lefebvre, Vivek Kwatra, and Greg Turk.
\newblock State of the art in example-based texture synthesis.
\newblock 2009.

\bibitem{wei2000fast}
Li-Yi Wei and Marc Levoy.
\newblock Fast texture synthesis using tree-structured vector quantization.
\newblock In {\em ACM SIGGRAPH}, 2000.

\bibitem{xu2020video}
Jingwei Xu, Huazhe Xu, Bingbing Ni, Xiaokang Yang, and Trevor Darrell.
\newblock Video prediction via example guidance.
\newblock {\em International Conference on Machine Learning (ICML)}, 2020.

\bibitem{ye2019compositional}
Yufei Ye, Maneesh Singh, Abhinav Gupta, and Shubham Tulsiani.
\newblock Compositional video prediction.
\newblock In {\em IEEE International Conference on Computer Vision (ICCV)},
  2019.

\bibitem{zhang2020vid2player}
Haotian Zhang, Cristobal Sciutto, Maneesh Agrawala, and Kayvon Fatahalian.
\newblock Vid2player: Controllable video sprites that behave and appear like
  professional tennis players.
\newblock {\em arXiv preprint arXiv:2008.04524}, 2020.

\bibitem{zhou2019dance}
Yipin Zhou, Zhaowen Wang, Chen Fang, Trung Bui, and Tamara~L Berg.
\newblock Dance dance generation: Motion transfer for internet videos.
\newblock {\em arXiv preprint arXiv:1904.00129}, 2019.

\bibitem{zhu2017unpaired}
Jun-Yan Zhu, Taesung Park, Phillip Isola, and Alexei~A Efros.
\newblock Unpaired image-to-image translation using cycle-consistent
  adversarial networks.
\newblock In {\em IEEE International Conference on Computer Vision (ICCV)},
  2017.

\end{thebibliography}
}
\section{Supplementary}

\noindent This supplementary section is organized as follows:
\begin{enumerate}
    \item Implementation details
    \item Unconditional Video Texture Baselines
    \item Video Quality Metric
    \item Unconditional Contrastive Audio-Video Textures
    \item Comparing Transition Probabilities
\end{enumerate}

\noindent Additionally, we include the following videos in \href{https://medhini.github.io/audio_video_textures/}{here}:
\begin{enumerate}
    \item An overview video explaining our method, results, comparisons to baselines for both unconditional and conditional settings, and videos for Figure 1, Figure 4 and MocoGAN videos.
    \item Additional results of comparisons to baselines for both conditional and unconditional settings. 
    \item Interpolation: Qualitative results of textures synthesized with and without interpolation. 
\end{enumerate}

\subsection{Implementation Details}

\noindent\textbf{Video Encoding.} We use the SlowFast~\citep{feichtenhofer2019slowfast} action recognition model pretrained on Kinetics-400~\citep{kay2017kinetics} for encoding the video segments. We divide the video into overlapping segments using a window of length 0.5 seconds and a stride of 0.2 seconds. Depending on the frame rate of the video, this yields segments with varying number of frames. Each of these segments is then encoded by SlowFast model into $\mathbb{R}^{512}$. Next the query and target are passed through two separate MLPs, each consisting of 3 linear layers interspersed with ReLU activations. The MLP maintains the size of the embedding such that the final outputs, $\phi(S)$ and $\psi(S)$ are in $\mathbb{R}^{512}$. We initialize the SlowFast model with weights pretrained on Kinetics dataset and fine tune the whole network end-to-end. We use the SGD optimizer with a learning rate of $1e$-$4$.

We also experimented with alternate video encoding networks, such as I3D (RGB + Flow) and ResNet3D (RGB + Flow). Both these networks performed on par with SlowFast for our task and we decided to use SlowFast for the final model as it does not require optical flow to be computed and can thus be trained end-to-end on the raw data.    

\noindent\textbf{Audio Encoding.}
We embed the audio segments using the VGGish model~\citep{hershey2017cnn} pretrained on AudioSet~\citep{gemmeke2017audio}. We remove the last fully connected layer from the model and use the output of the final convolutional layer as audio features. The learned audio representations for the source audio segments $\varphi(A^c)$ and the conditioning audio segments $\varphi(A^c)$ are in $\mathbb{R}^{128}$.
 
\noindent\textbf{Interpolation.} We typically set the number of interpolated frames to be added to be 4. This increases the FPS of the synthesized video by a factor of 3 (i.e. 2 frames is converted to 4). When there is no jump, the frames are repeated 3 times, to ensure the overall FPS of the video is the same.   

\noindent\textbf{Temperature tuning and threshold.}
For training our Contrastive Video Texture model, we experimented with multiple values of temperature ($\tau$) and found 0.1 to work the best. At test time, setting the temperature to 0.1 and threshold ($t$) to 100.0\% regurgitates the original video, as the segment with the highest probability (\ie the positive segment) is always chosen and all values below that are thresholded to 0. Increasing the temperature and decreasing the threshold increases the entropy and allows for more random transitions in the output. We found that the number of transitions vs the temperature is fairly constant across all videos and include details in Sec.~\ref{sec:tp}. Upon manually analyzing several synthesized textures, we found that temperatures in the range of $0.3 - 0.7$ and thresholds in the range of $99.8 - 99.98$ are optimal for synthesizing videos which are temporally smooth yet different from the original video. We synthesized video textures at different temperatures and observed that our model is capable of synthesizing multiple plausible output videos given a single input video. 

In case of the Classic algorithm and its variants, we found that the temperature and threshold were not generalizable across videos and only a specifc temperature (sigma) and threshold value resulted in temporally coherent textures. For the classic methods, the right hyperparameter values had to be chosen on a per video basis, by manually analyzing the synthesized textures over a range of hyperparameter values, which was quite tedious.  Thus, ``learning'' feature representations and the distance metric as opposed to computing distances on raw pixels helped ease the task of choosing the right hyperparameters. 

\noindent\textbf{Combining $T_a$ and $T_v$.} Smaller values of $\alpha$ allow for better audio-video synchronization but at the cost of continuity in the video. For most results reported here, we set $\alpha$ to either 0.5 or 0.7. 
\subsection{Unconditional Video Textures: Baselines}
\label{sec:uncondbaselines}

We first provide an overview of the classic video texture algorithm introduced in~\cite{schodl2000video} followed by the descriptions of the baselines. 

\medskip
\noindent\textbf{$\bullet$ Classic Video Textures:} The classic video textures algorithm proposed in~\cite{schodl2000video} computes a distance matrix $D$ of pairwise distances between all frames in the video. The distance is computed as the L2-norm of the difference in RGB values between pairs of frames. Next, the distance matrix $D$ is filtered with a 2 or 4-tap filter with binomial weights to produce matrix $D'$. The stride used while filtering is 1. If the input video is short, oftentimes this approach would not be able to find good transitions from the last frame and reaches a dead end. To avoid this, they use Q-learning to predict the anticipated (increased) ``future cost'' of choosing a given transition, given the future transitions that such a move might necessitate. This gives rise to $D''$. The transition probabilities $P''$ are computed from $D''$ as $P''_{i,j} = \textrm{exp}(-D''_{i+1,j}/\sigma)$.

To synthesize a texture, a frame $i$ is chosen at random. This is added to the output sequence of frames. After displaying frame $i$, the next frame $j$ is selected according to $P_{i,j}$. To improve the quality of the textures and to suppress non-optimal transitions they adopt a two step pruning strategy. First, they choose the optimal transition with the maximum transition probability, next they set all probabilities below some threshold to zero and pick a random transition from the non-zero probabilities. The output sequence is generated one frame at a time. 

We generate textures using the algorithm above. Following the convention in~\cite{schodl2000video}, we set sigma to be a small multiple of the average (non-zero) values in the distance matrix. We tune this small multiple and the threshold on the train set and use the same values on the test set. 

For an apples-to-apples comparison, we fix some of the shortcomings of the classic algorithm and compare to these modified versions described below. 

\noindent\textbf{$\bullet$ Classic+} During inference, the number of frames appended to the output texture is the stride with which the initial video was segmented. While this stride is 1 for the classic algorithm, it is greater than 1 for our contrastive method. To ensure the difference in perceptual quality isn't due to just the changes in stride length, we modify the classic algorithm to increase the stride during inference to be the same as our contrastive model. The distance matrix is still computed pairwise between frames but instead of appending a single frame, we append stride number of frames to the output. This stride is set to be the same value as our contrastive model. 

\noindent\textbf{$\bullet$ Classic++.} To further reduce the gap between classic+ and contrastive method, we apply a stride > 1 while filtering the distance matrix $D$ with the tap filter. This is equivalent to the approach we use in contrastive, which is dividing the video into overlapping segments  of window $W$ and stride $s$. 

\noindent\textbf{$\bullet$ Deep Classic:} Additionally, we also tried replacing the frame-wise features in the classic algorithm with learned representations from a pre-trained resnet. 

\subsection{Video Quality Metric}

\begin{figure*}[h]
    \centering
    \includegraphics[width=0.9\textwidth]{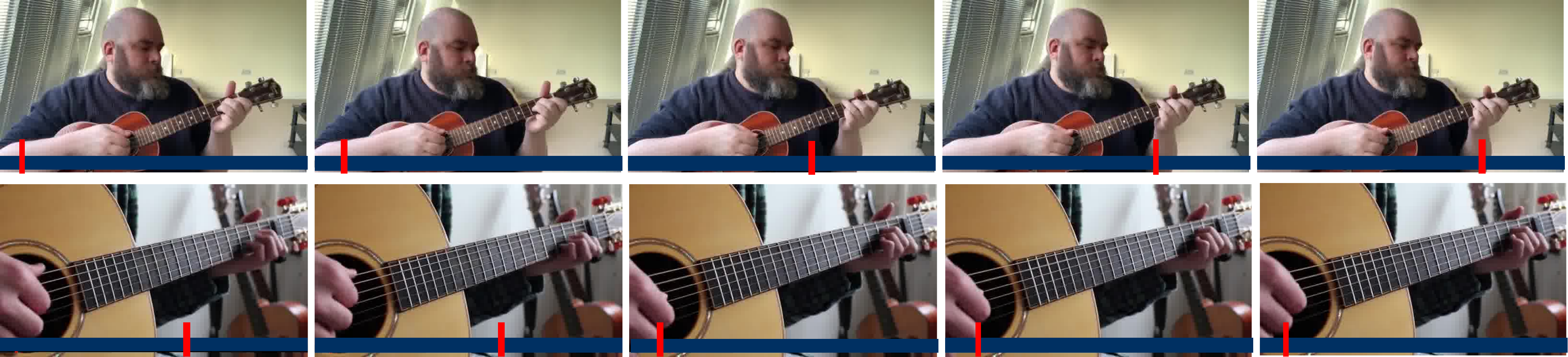}
    \caption{The figure shows frames from two different videos synthesized by our method. Red bar indicates position of the original video being played. The transition happens at the third frame and is seamless in both cases. The first is a forward jump and the second is a backward jump.}
    \label{fig:trans}
\end{figure*}

We report the average FVD~\citep{unterthiner2018towards} computed between the original videos and the synthesized video textures for both the unconditional and conditional settings. As shown in Tab.~\ref{tab:uncond_fvd}, our Contrastive method obtains the lowest FVD of  and is the best performing method. 

For the conditional setting, both Random Clip and Audio NN baselines involve extracting and replaying a portion of the original video, and hence receive the lowest FVD scores. Our method performs better than Classic+Audio and VRB. 


\begin{table}[t]
    \centering
    \begin{tabular}{ll}\toprule
    {\bf Method}  & {\bf FVD $\downarrow$}
    \\ \midrule
    Classic & 379 \\
    Classic Deep & 443\\
    Classic+ & 208\\
    Classic++ & 226 \\\midrule
    Contrastive & \textbf{151}\\\bottomrule
    \end{tabular}
    \caption{\textbf{Unconditional Video Texture Synthesis.} We report FVD scores for all the baselines and our method. A lower FVD score is better.}
    \label{tab:uncond_fvd}
\end{table}

\begin{table}[t]
    \centering
    \begin{tabular}{ll}\toprule
    {\bf Method}  & {\bf FVD $\downarrow$}
    \\ \midrule
    Classic+Audio & 536 \\
    Random Clip & \textbf{135} \\
    Audio NN & \textbf{138} \\
    VRB & 415 \\\midrule
    Contrastive & \textbf{158}\\\bottomrule
    \end{tabular}
    \caption{\textbf{Audio-Conditioned Video Textures.} We report FVD scores for all the baselines and our method. A lower score is better.}
    \label{tab:cond_fvd}
\end{table}

\subsection{Unconditional Contrastive Audio-Video Textures}

In this variant of our method, we combined audio features with the 3D video features to train our contrastive model. Assuming the input video has corresponding audio, we extract overlapping audio segments following the same approach used for video segments in Sec. 3 of the main paper. Next, we use two separate encoder heads for the query and target audio pairs. We then concantenate the audio encodings with the 3D video encodings and pass the fused features through a linear layer. The network is trained using contrastive learning, using Eq. 2 of the main paper. We observed that the resulting audio-video textures were smooth not only in the video domain but the audio domain as well. However, they were not as diverse as the ones trained with just video features. For this reason we used the video only model.    

\subsection{Comparing Transition Probabilities}

\begin{figure}[h]
    \centering
    \includegraphics[width=\linewidth]{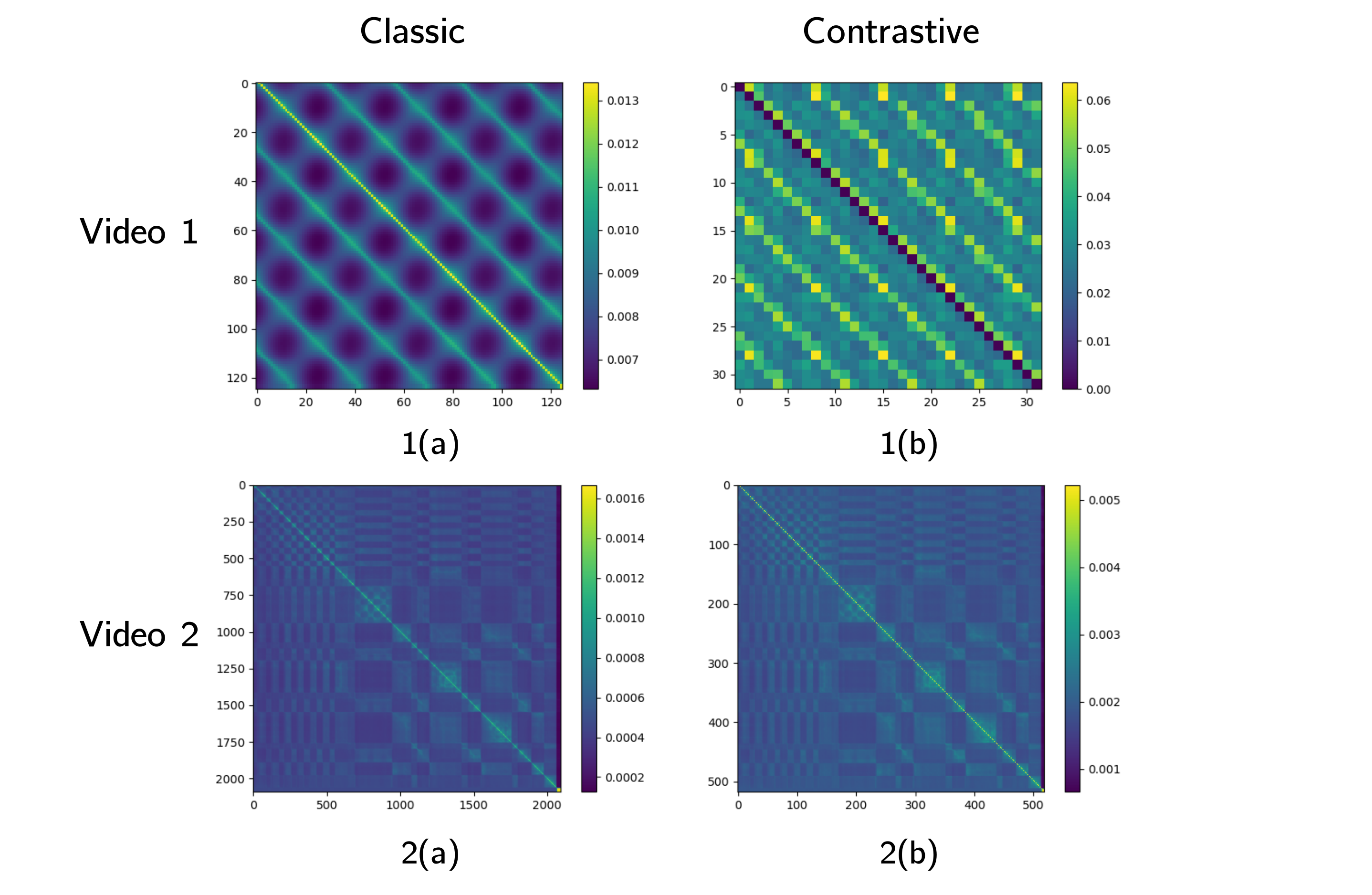}
        \caption{Transition probability matrix for two different videos (in each row) for both classic and contrastive methods. }
        \label{fig:tranprob}
\end{figure}

\begin{figure}
        \centering
        \includegraphics[width=\linewidth]{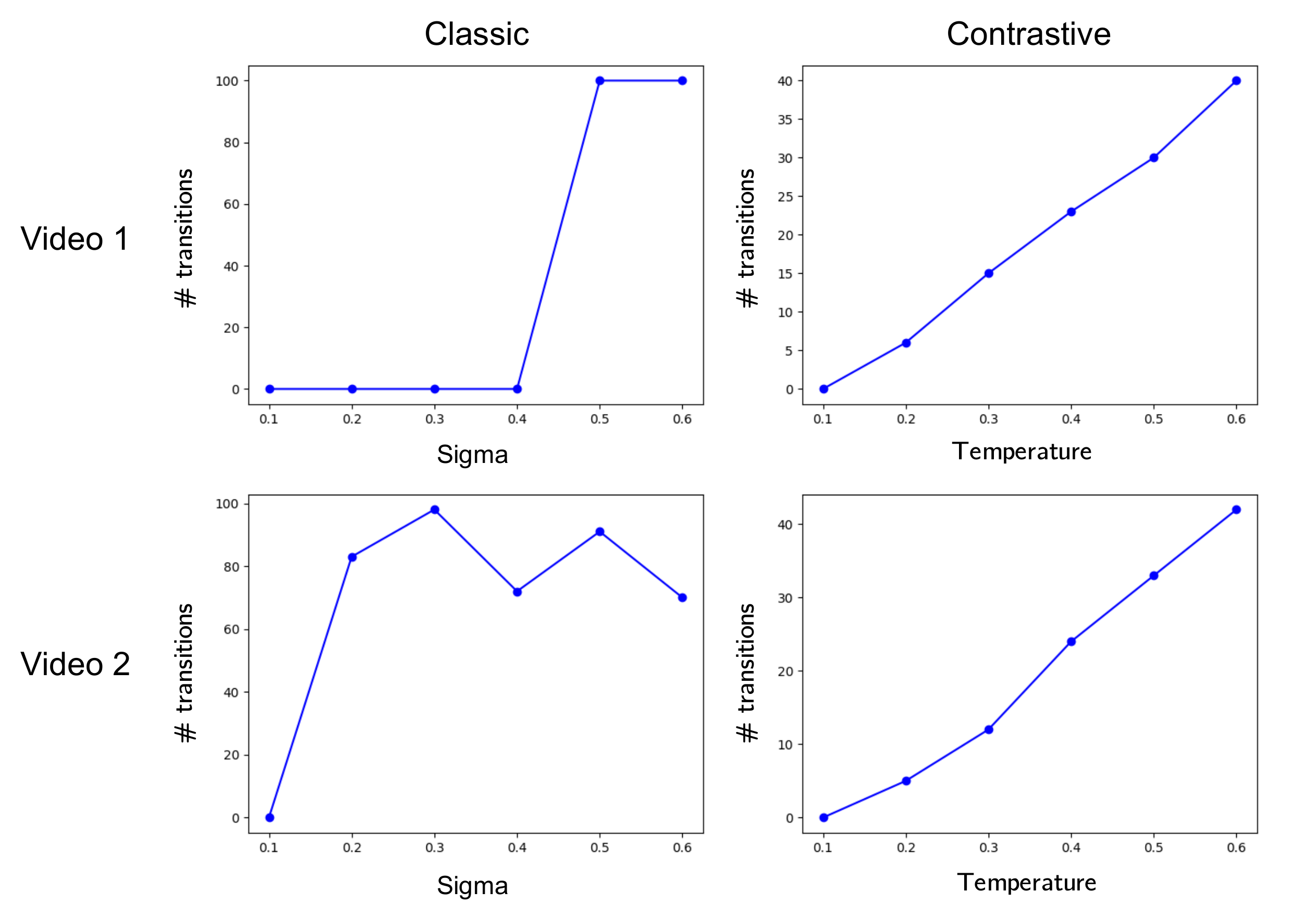}
        \caption{Number of transitions vs Sigma for Classic and Number of transitions vs Temperature for Contrastive.}
        \label{fig:jumps}
\end{figure}

\label{sec:tp}
We compare the \textbf{transition probability matrices} generated by both classic and contrastive methods. Fig.~\ref{fig:tranprob} shows the transition probability matrices for two different videos generated by Classic (1a, 2a) and Contrastive (1b, 2b) methods. It can be observed from the diagonal lines in the figure that the classic method assigns the same value to multiple frames whereas our method picks up on subtle differences and assigns different scores. This emphasizes that the distance metric learned by our method is better at distinguishing frames. 

Fig.~\ref{fig:jumps} shows the variation in the number of transitions with sigma for the classic technique and with temperature for the contrastive technique. 
Number of transitions increases linearly with temperature for contrastive method whereas for the classic technique we found no such correlation. Moreover, a temperature of 0.3 and a threshold of 0.01 results in ~15-20 jumps across all videos. There was no such strong correlation for the classic technique, making it necessary to tune hyperparameters on a per video basis.

Fig.~\ref{fig:trans} shows some transitions in the video textures generated by contrastive model.

\end{document}